\documentclass[sigconf]{acmart}
\AtBeginDocument{%
  \providecommand\BibTeX{{%
    \normalfont B\kern-0.5em{\scshape i\kern-0.25em b}\kern-0.8em\TeX}}}

\setcopyright{acmcopyright}
\copyrightyear{2023}
\acmYear{2023}
\setcopyright{acmlicensed}\acmConference[WWW '23]{Proceedings of the ACM Web Conference 2023}{May 1--5, 2023}{Austin, TX, USA}
\acmBooktitle{Proceedings of the ACM Web Conference 2023 (WWW '23), May 1--5, 2023, Austin, TX, USA}
\acmPrice{15.00}
\acmDOI{10.1145/3543507.3583399}
\acmISBN{978-1-4503-9416-1/23/04}

\usepackage{xcolor}
\usepackage{colortbl}
\usepackage{graphicx}
\usepackage{booktabs}       
\usepackage{amsfonts}       
\usepackage{nicefrac}       
\usepackage{microtype}      
\usepackage{subfigure}
\usepackage{amsthm}
\usepackage{algorithm}
\usepackage{algorithmic}
\usepackage{multirow}
\usepackage{enumitem}
\usepackage{callouts}

\usepackage{amsmath,amsfonts,bm}

\def\eqref#1{equation~\ref{#1}}

\def\1{\bm{1}}

\def\vu{{\bm{u}}}
\def\vv{{\bm{v}}}

\def\vx{{\bm{x}}}

\def\mA{{\bm{A}}}

\def\mD{{\bm{D}}}
\def\mE{{\bm{E}}}

\def\mI{{\bm{I}}}

\def\mL{{\bm{L}}}
\def\mM{{\bm{M}}}

\def\mQ{{\bm{Q}}}

\def\mU{{\bm{U}}}
\def\mV{{\bm{V}}}

\def\mX{{\bm{X}}}
\def\mY{{\bm{Y}}}
\def\mZ{{\bm{Z}}}

\DeclareMathAlphabet{\mathsfit}{\encodingdefault}{\sfdefault}{m}{sl}
\SetMathAlphabet{\mathsfit}{bold}{\encodingdefault}{\sfdefault}{bx}{n}

\def\gD{{\mathcal{D}}}
\def\gE{{\mathcal{E}}}

\def\gG{{\mathcal{G}}}

\def\gL{{\mathcal{L}}}

\def\gN{{\mathcal{N}}}

\def\gT{{\mathcal{T}}}

\def\gV{{\mathcal{V}}}
\def\gW{{\mathcal{W}}}

\newcommand{\R}{\mathbb{R}}

\DeclareMathOperator*{\argmin}{arg\,min}

\DeclareMathOperator{\sign}{sign}

\def\vnu{{\bm{\nu}}}

\newcommand{\wtW}{{\boldsymbol{\mathcal{W}}}}

\newcommand{\wtV}{\widetilde{\boldsymbol{{V}}}}
\newcommand{\iprod}[2]{\langle #1,\,#2 \rangle}
\newcommand{\norm}[1]{{|\kern-1.125pt|} #1 {|\kern-1.125pt|}}

\newcounter{bxincomm}
\definecolor{aqua}{rgb}{0.00,0.67,0.80}

\newcounter{ygcounter}

\newcommand{\ygc}[1]{\ygc{\stepcounter{ygcounter}{\bf [YG's comment \arabic{ygcounter}: #1]}\;}}

\begin{document}
\title[Robust Graph Representation Learning for Local Corruption Recovery]{Robust Graph Representation Learning for Local Corruption Recovery}

\settopmatter{authorsperrow=4}
\author{Bingxin Zhou}
\authornote{Both authors contributed equally to this research.}
\affiliation{%
  \institution{Shanghai Jiao Tong\\University}
  \state{Shanghai}
  \country{China}}
\affiliation{%
  \institution{The University of Sydney}
  \state{NSW}
  \country{Australia}}

\author{Yuanhong Jiang}
\authornotemark[1]
\affiliation{%
  \institution{Shanghai Jiao Tong\\University}
  \state{Shanghai}
  \country{China}}
\author{Yu Guang Wang}
\affiliation{%
  \institution{Shanghai Jiao Tong\\University}
  \state{Shanghai}
  \country{China}}
\author{Jingwei Liang}
\affiliation{%
  \institution{Shanghai Jiao Tong\\University}
  \state{Shanghai}
  \country{China}}
\author{Junbin Gao}
\affiliation{%
  \institution{The University of Sydney}
  \state{NSW}
  \country{Australia}}
\author{Shirui Pan}
\affiliation{%
  \institution{Griffith University}
  \state{QLD}
  \country{Australia}}
\author{Xiaoqun Zhang}
\affiliation{%
  \institution{Shanghai Jiao Tong University}
  \state{Shanghai}
  \country{China}}


\renewcommand{\shortauthors}{Zhou and Jiang, et al.}

\begin{abstract}
  The performance of graph representation learning is affected by the quality of graph input. While existing research usually pursues a globally smoothed graph embedding, we believe the rarely observed anomalies are as well harmful to an accurate prediction. This work establishes a graph learning scheme that automatically detects (locally) corrupted feature attributes and recovers robust embedding for prediction tasks. The detection operation leverages a graph autoencoder, which does not make any assumptions about the distribution of the local corruptions. It pinpoints the positions of the anomalous node attributes in an unbiased mask matrix, where robust estimations are recovered with sparsity promoting regularizer. The optimizer approaches a new embedding that is sparse in the framelet domain and conditionally close to input observations. Extensive experiments are provided to validate our proposed model can recover a robust graph representation from black-box poisoning and achieve excellent performance.
\end{abstract}




\maketitle

\section{Introduction}
\label{sec:intro}
Graph neural networks (GNNs) have received tremendous success in the past few years \cite{bronstein2017geometric,wu2020comprehensive,zhang2020deep,zhou2020graph}. Graphs, as the input of GNNs, record useful features and structural information. They exist widely in many fields, such as biomedical science \cite{ahmedt2021graph}, social networks \cite{fan2019graph}, and recommender systems \cite{wu2020graph}.

Similar to other types of real-world data, inaccurate observations are ubiquitous in graphs with a noticeable side-effect for graph representation learning. For instance, fraudulent users in social media tend to fake user avatars or online activities. A recommender might be dysfunctional by mislabeled items or users. Such disruptive observations hinder the model fitting and prediction. The feature aggregation in graph representation learning accentuates the negative influence of irregular entities on their neighborhoods, resulting in a misleading latent feature representation for the predictor.

\begin{figure*}
    \centering
    \includegraphics[width=\textwidth,trim={3mm 0 0 0},clip]{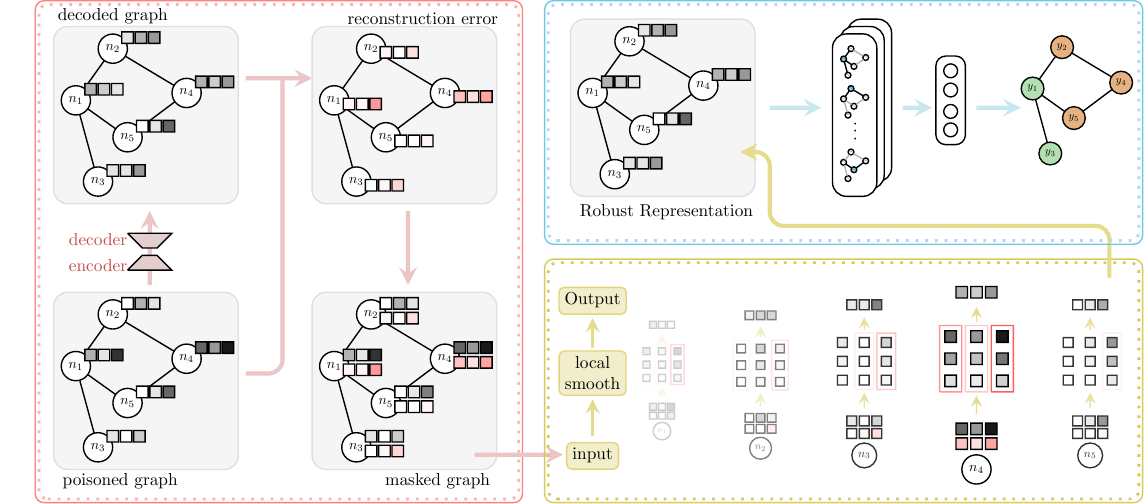}
    \caption{Illustrative architecture of the proposed \textsc{MAGnet}. Input feature attributes (\textcolor{black!60!}{gray}) are assumed locally corrupted. The first module constructs a mask matrix (\textcolor{red!80!black!50!}{red}) as the indices of locally-corrupted input feature attributes (\textcolor{black!60!}{gray}). They are sent to an inertia ADMM optimizer (\textcolor{yellow!80!black!70!}{yellow}) to iterate a robust feature representation, which is functioned as hidden embeddings by typical graph convolutional layers. It can be encoded by other convolutions or sent to a predictor (\textcolor{cyan!80!black!70!}{blue}). }
    \label{fig:main}
\end{figure*}

Existing works are aware of the harmful graph anomalies. i) Graph anomaly detection \cite{ding2019deep,ma2021comprehensive,peng2020deep,zhu2020anomaly} identifies the small portion of problematic nodes, such as fraudulent users in a social network graph; ii) graph defense\cite{dai2018adversarial,xu2020adversarial,zugner2019adversarial} refines the learning manner of a classifier to provide promising predictions against potential misleading entities that hurt the model training; iii) optimization-based graph convolutions smooth out global noise, e.g., observation errors in data collection, by special regularization designs \cite{chen2021graph,liu2021elastic,zhou2021admm,zhu2021interpreting}. However, the first approach detects the whole irregular node rather than individual node attributes, and it does not amend the flawed representation of the identified outlier. The second approach provides an adequate solution to protect the prediction performance, but most researches focus on graph rewiring, as edges are believed more vulnerable to adversarial attacks. The third choice, although operates on node attributes, only makes implicit optimization against local corruptions. They usually assume that feature corruptions are normally observed in all inputs. 

In this paper, we propose an alternative approach by developing a `\textit{detect-and-then-recover}' strategy to protect graph representation learning from a small portion of hidden corruptions in the input node attributes. In particular, an unsupervised encoder module first detects suspicious attributes and assigns a mask matrix to expose their positions. The detector requires no prior knowledge of the distribution of the anomalous attributes. The constructed mask guides a robust reconstruction of the initial input with sparsity promoting regularizers. 
To guarantee an expressive representation that simultaneously reconstructs irregular attributes, preserves small-energy patterns, and eliminates global noise, the regularization is operated on multi-scale framelet coefficients \cite{dong2017sparse,zheng2021framelets}. Meanwhile, the posterior mask guarantees that the signal recovery is predominantly conducted at the essential (anomalous) spots. The optimized embedding is updated iteratively, which acts similarly to a graph convolutional layer that constantly smooths the hidden feature representation for GNN prediction tasks. 

Combining the three key ingredients of \underline{\textbf{M}}ask, \underline{\textbf{A}}DMM, and \underline{\textbf{G}}raph, we name our model as \textsc{MAGnet}. Figure~\ref{fig:main} illustrates the three components of unsupervised mask construction, localized robust optimization, and graph representation learning. \textsc{MAGnet} detects local corruptions in an unsupervised manner and approximates a robust graph representation. As indicated above, the alternating direction method of multipliers (ADMM) algorithm \cite{gabay1976dual} is adopted in our optimization procedure. 


\section{Problem Formulation}
\label{sec:formulation}
We use $\gG=(\gV,\gE,\mX)$ to denote an undirected attributed graph with $n=|\gV|$ nodes and $|\gE|$ edges, where the latter is usually described by an adjacency matrix $\mA\in \R^{n\times n}$. The observed $d$-dimensional node features $\mX\in\R^{n\times d}$ is a noised version of the ground truth signal $\bar{\mU}$, which reads $\mX=\bar{\mU}+\mE_1+\mE_2$
We call $\mE_1$ global noise and $\mE_2$ outliers or local corruptions. The main difference between $\mE_1$ and $\mE_2$ is that the former might universally exist on the entire graph, and the latter generally takes a small chance of existence so it would not be observed from a large set of nodes or node attributes. We hereby make the following assumptions on the properties of $\bar{\mU}, \mE_1$ and $\mE_2$: 
\begin{enumerate}[label=(\alph*),leftmargin=*]
  \item The global noise $\mE_1$ follows some distribution $\gD$ with an expectation of $0$;
  \item The outliers take a small portion of the entire graph, or the $\mE_2$ is a sparse matrix; and
  \item The observations $\mX$ are close to $\bar{\mU}$ at regular locations.
\end{enumerate}
The associated objective function to estimate $\bar{\mU}$ reads
\begin{equation}
    \label{eq:objective_init}
    \begin{aligned}
        \min_{\mU} ~ \alpha {\rm Reg}(\mU) + {\rm Loss}(\mE_2) 
        \quad \mathrm{s.t.}\quad ~ \mX|_{\mM}=(\mU+\mE_1+\mE_2)|_{\mM},
    \end{aligned}
\end{equation}
where $\alpha$ is tunable, and $\mM$ is a mask matrix of outliers indices. The ${\rm Reg}(\cdot)$ and ${\rm Loss}(\cdot)$ denote the regularizer and loss function satisfying assumptions (a) and (b), respectively.

\begin{itemize}[leftmargin=*]
  \item \textbf{Choice of ${\rm Loss}(\mE_2)$: } The loss function (\ref{eq:objective_init}) finds a robust approximation $\mU$ from the input feature $\mX$. As required by assumption (b), a measure with respect to the sparsity of $\mE_2$ has to be optimized. The best choice for sparsity, in theory, is $\ell_0$-norm which counts the number of nonzero entries in $\mE_2$. Unfortunately, the $\ell_0$ constrained problem is NP-hard. In substitute, the $\ell_1$-norm allows feasible solution for (\ref{eq:objective_init}) that promotes a good sparsity measure for $\mE_2$ \cite{chen2015signal}. Also \cite{li2017radar,peng2018anomalous} implemented residual analysis-based \emph{anomaly detection} methods for labeling suspicious outliers. 
  \item \textbf{Choice of ${\rm Reg}(\mU)$: } The regularization on $\mU$ is a penalty complementary to the loss function, which controls the noise level of $\mX$ by the smoothness of $\mU$, and it is usually quantified by some energy measurements. For instance, GCN \cite{Kipf2016semi} utilizes normalized Dirichlet energy of $\mU$ by ${\rm tr}(\mU^{\top}\tilde{\mL}\mU)$, where $\tilde{\mL}$ denotes the normalized graph Laplacian from $\tilde{\mL}=\mI-\mD^{-\frac12}\mA\mD^{-\frac12}$ with the adjacency matrix $\mA$ and its degree matrix $\mD$. Minimizing such energy encourages message transmissions among connected nodes. By extracting the summary of neighborhood space, unnoticeable noise is likely to be smoothed out. In addition, minimizing the Dirichlet energy implies a low-rank smooth solution to $\mU$ \cite{monti2017geometric}. Such a \emph{graph smoothing effect} \cite{chen2021graph,liu2021elastic,zhu2021interpreting} has been implemented in many spatial-based graph convolution designs \cite{klicpera2018predict,wu2019simplifying,xu2018representation}. Alternatively, spectral-based methods regularize the graph signal in a transformed domain by $\mL$. For example, Dong \cite{dong2017sparse} and Zhou \textit{et al.} \cite{zhou2021admm} minimized the $\ell_1$-norm total variation of framelet coefficients, and Mahmood \textit{et al.} \cite{mahmood2018adaptive} considered a similar regularization in the wavelet domain. As the total variation reflects redundant local fluctuations, a solution to minimize it is believed to remove noise and preserve local patterns at the same time.

\item \textbf{Restriction on $\mE_1+\mE_2$: }
Compared to the main ingredients of the objective function, the treatment of the fidelity constraint is rather trivial when the outliers $\mE_2$ do not exist. Usually a regularizer is adopted to minimize the difference between $\mX$ and $\bar{\mU}+\mE_1$, i.e., $\mX-\mU$. However, when $\mE_2$ is assumed to exist at a small proportion, it is undesired to force $\mU$ to approximate $\mX$ especially on anomalous locations. Instead, a conditional approximation is placed with the index matrix $\mM$, in which case only attributes at regular positions are required aligned. The target is then $\min{\rm Reg}\left(\mM\odot(\mX-\mU)\right)$ with some penalty ${\rm Reg}(\cdot)$, and it can be appended to the main objective by
\begin{equation} \label{eq:objective_init_2}
    \min_{\mU} \alpha {\rm Reg}(\mU) + {\rm Loss}\left(\mM\odot(\mX-\mU)\right).
\end{equation}
The above optimization for graph signal recovery is well-defined. Nevertheless, it has three issues in practice. First, a direct minimization of the Dirichlet energy in the spatial domain usually falls into the pitfall of over-smoothing, where the recovered graph loses expressivity drastically \cite{balcilar2020analyzing}. On the other hand, spectral transforms can be sophisticated and time-consuming, which is generally circumvented by the majority of studies. Second, fixing the $\ell_1$ or $\ell_2$ norm can be too restricted for an arbitrary dataset or application, which could recover a less robust graph representation that affects the prediction task later on. Last but not least, attaining the mask matrix $\mM$ in (\ref{eq:objective_init_2}) can be nasty in practice, as the prior knowledge of $\mM$ is generally inaccessible.
\end{itemize}

In order to best handle the identified problems, we propose an objective function design that combines masked $\ell_q$ ($1\leq q\leq 2$) reconstruction error term and $\ell_p$ ($0\leq p\leq 1$) regularization, i.e.,
\begin{equation}\label{eq:objective_intro}
    \min_{\mU} \|\vnu\boldsymbol{\gW}\mU\|_{p,G}+\frac12\|\mM\odot(\mU-\mX)\|^q_{q,G},
\end{equation}
which is convex at $p=1$ and non-convex when $0\leq p<1$, regardless of $q$. Equation~(\ref{eq:objective_intro}) enforces the sparse representation by a \textit{graph framelet system }\cite{dong2017sparse,wang2019tight,zheng2021framelets,zheng2022decimated}. Through framelet transforms, an input graph signal is decomposed into a low-pass coefficient matrix and multiple high-pass coefficient matrices. The former includes general global information about the graph, and the latter portrays the local properties of the graph at different degrees of detail. 

Specifically, the decomposition is realized by a set of multi-scale and multi-level framelet decomposition operators $\boldsymbol{\gW}$ under the framelet system, with each $\boldsymbol{\gW}_{k,l}$ be an orthonormal basis at $(k,l) \in \{(0, J)\}\cup\{(1,1), \dots,(1, J), \dots(K, 1), \dots,(K, J)\}$. This decomposition operator forms the spectral representation of $\mU$, where its global trend is reflected in the low-pass coefficients $\boldsymbol{\gW}_{0, J}\mU$. Meanwhile, the high-pass coefficients $\boldsymbol{\gW}_{k,l}\mU$ records detailed information at scale $k$ and level $l$ ($1\leq k\leq K, 1\leq l\leq J$). They reflect local patterns or noises of the signal. The coefficients at a higher level $l$ contain more localized information with smaller energy.

In addition, we replace $\alpha$ in (\ref{eq:objective_init_2}) with a set of hyper-parameters $\vnu$ to normalize the high-frequency framelet coefficients $\boldsymbol{\gW}\mU$ of the approximated $\mU$ in different scales. For example, an initialization $\nu_0$ defines $\vnu_{0,J}=0$ for the low-pass coefficients and $\vnu_{k,l}=4^{-l-1}\nu_0$ for high-pass coefficients. 

The \textit{signal reconstruction error} $\|\mM\odot(\mU-\mX)\|_q$ guarantees the optimal representation $\mU$ to be conditionally close to the masked feature matrix $\mX$ of the given graph. We generalize the conventional $\ell_2$ error to a $\ell_q$ penalty ($1\leq q\leq 2$). Meanwhile, the $\ell_p$ regularization measures the graph total variation \cite{sandryhaila2014discrete} in the framelet domain, where pursuing a small total variation recovers the graph signal from corrupted attribute noises. When $p$ approaches $1$ from $2$, the framelet representation $\boldsymbol{\gW}\mU$ gains increasing sparsity. 

We treat $p,q$ as tunable parameters to allow sufficient freedom in our design for selecting the most qualified optimization model that fits the data and training requirements. Furthermore, we replace the ordinary Euclidean $\ell_k$-norm with a graph $\ell_k$-norm, denoted as $\ell_{k,G}$, to assign higher penalties to influential high-degree nodes. For an arbitrary node $\vv_i$ of degree $\mD_{ii}$, $\|\vv_i\|_{k,G}:=\left(\|\vv_i\|^k \cdot\mD_{ii}\right)^{\frac1k}$.

Compared to the initial design in (\ref{eq:objective_init}), (\ref{eq:objective_intro}) made three adjustments to tackle the identified issues.
First, minimizing the first component smooths out $\mE_1$ under the assumption (a) 
from high-pass framelet coefficients, which avoids information loss by spatial Dirichlet energy minimization. In other words, the global noise $\mE_1$ is removed without sacrificing small-energy features. Secondly, we adopt the values of $p,q$ to adaptively restrict the sparsity of recovered graph with $p\in[1,2], q\in[0,1]$. As introduced in Section~\ref{sec:admm}, the optimization can be solved by an inertial version of the alternating direction method of multipliers with promising convergence.

A potential solution to the unreachable mask matrix $\mM$ is to add a sub-problem of optimizing (\ref{eq:objective_intro}), which introduces a two-stage optimization problem. However, this approach blows up the difficulty of solving the existing problem and the mask region could be too complicated to reveal. Instead, we consider an unsupervised GNN scheme to automate the discovery of anomalous positions. We approximate the anomaly score of $\mX$ with a classic graph anomaly detection scheme that looks for the reconstruction error between the observed and reconstructed matrices from neural networks. 

\section{Mask Matrix Generation with Graph Anomaly Detection}
\label{sec:gae}
Graph anomaly detection is an important subproblem of the general graph signal recovery where outliers are assumed to exist in the input. A detector identifies nodes with $\mE_2$ as \emph{community anomalies} \cite{ma2021comprehensive}, which are defined as nodes that have distinct attribute values compared to their neighbors of the same community. The underlying assumption here is that the outlier is sparse and their coefficients are antipathetic from the smooth graph signal $\mU$. We hereby consider GNN-based algorithms to investigate the difference between each node from the representation of its local community.

\subsection{Graph Autoencoder}
Autoencoder is a powerful tool for reconstructing corrupted objects \cite{aggarwal2017linear,kovenko2020comprehensive}. GAE \cite{kipf2016variational} is a GCN-based autoencoder, which has become a frequent solution in graph anomaly detection \cite{ding2019deep,peng2020deep,zhu2020anomaly}. A GAE layer takes the information of $\gG$ to obtain an embedding $\mZ\in\R^{n\times h}$ of a hidden size $h$ by $\mZ = {\rm GCN}(\mX,\mA)$. When recovering the graph attributes from its hidden embedding, a feed-forward network is trained to find $\mX^{\prime}={\rm FFN}(\mZ)$, a smoothed version of $\mX$ that is believed to remove essential noise or minor corruptions of the raw input $\mX$. In \cite{ding2019deep,peng2018anomalous} for graph anomaly detection, the learning objective of the neural network is to best reconstruct the graph attribute $\mX^{\prime}$, which loss function is the weighted average of the squared \emph{approximation errors}, i.e., $ \gL=\|\mX-\mX^{\prime}\|_2$.

Once the graph is reconstructed, it is straightforward to find outliers by comparing the input and output representation. For node entities, a corrupted object has different patterns from its local community, so its initial representation should be distinct from the smoothed representation. The approximation error is thus an appropriate indicator for diagnosing such divergence. For a particular node $\vv_i$, we calculate ${\rm score}(\vv_i)=\|\vx_i-\vx_i^{\prime}\|_2$.

While the outliers exist rarely, we adopt $\ell_2$ penalty for approximating the approximation error (and anomalous scores) for two reasons. First, reconstructing the noisy signal by an autoencoder aims at finding an overall optimal representation without extreme errors. In this sense, the MSE loss is more sensitive to large errors than the MAE loss. Meanwhile, as we do not assume prior knowledge of the distribution of data noise, Gaussian noise (in alignment with $\ell_2$ penalty) is more general among other choices (such as the Laplacian distribution with $\ell_1$ penalty). On the other hand, the sparsity requirement is satisfied in the later stages of our implementation. As introduced in the next section, the mask matrix construction is instructed by an MAE-loss, which favors a sparse distribution of the local noise in the attribute matrix.

\begin{figure}[t]
    \centering
    \includegraphics[width=0.8\columnwidth]{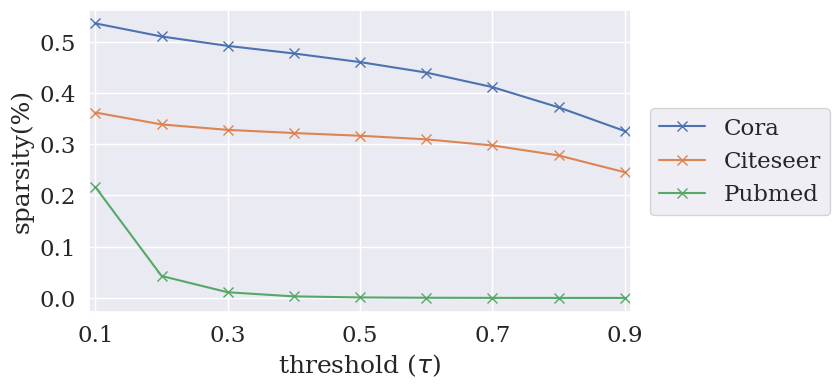}
    \caption{The sparsity of GAE-approximated $1-\mM$ is directly controlled by the threshold $\tau$.}
    \label{fig:sparsity_gae}
\end{figure}

\subsection{Mask Matrix Generation}
Our primary focus is on graph recovery without explicitly considering the disordered node connections or the adjacency matrix. Also, unlike conventional anomaly detection tasks that locate the problematic nodes, we aim at generating a mask matrix of size $n\times d$ that identifies the corrupted node attributes, and it is defined as
\begin{equation}
\label{eq:M}
    \mM=1-{\rm threshold}(\|\mX-\mX^{\prime}\|_1,\tau).
\end{equation}
The established $\mM$ approximates the element-wise errors by differentiating the raw and the recovered feature matrices from a trained graph autoencoder. Optionally, $\mM$ can be binarized with a tunable threshold $\tau\in(0,1)$. Consequently, $\mM_{ij}=1$ suggests assigning a new value to the $j$th feature of node $i$ to avoid potential corruption.

Under assumption (b) that $\mE_2$ exists rarely in node attributes, the majority of the mask matrix should be $1$, i.e., $1-\mM$ is a sparse matrix. This requirement is guaranteed by both the autoencoder algorithm and the thresholding function. Because the encoder smoothes $\mU$ to approach $\mX$ and the magnitude of $\mE_1$ is assumed smaller than $\mE_2$'s, the difference between $\mX$ and $\mX^{\prime}$ mainly comes from $\mE_2$. On top of that, the tunable $\tau$ controls the sparsity of $\mM$. 
Figure~\ref{fig:sparsity_gae} provides an empirical example with three citation networks that are locally perturbed by the attribute injection \cite{ding2019deep}. A large $\tau$ reduces the number of non-zero elements in $1-\mM$ from a considerably sparse level to a lower degree.

\section{Optimization algorithm}
\label{sec:admm}
This section describes the optimization algorithm to solve (\ref{eq:objective_intro}). As described in (\ref{eq:iadmm}), it is an inertial version of the celebrated alternating direction method of multipliers (ADMM) \cite{gabay1976dual}.

\subsection{An inertial ADMM}
Denote $\mZ = \wtW\mU$, then we can rewrite (\ref{eq:objective_intro}) as
\begin{equation}
    \label{eq:objective_node_full_Z}
    \begin{aligned}
        \min_{\mU,\mZ} &~\|\vnu\mZ\|_{p,G}+\tfrac12\|\mM\odot(\mU-\mX)\|^q_{q,G}, \quad
        \mathrm{s.t.} & \mZ = \wtW\mU.
    \end{aligned}
\end{equation}
This forms a standard formulation of problems that can be solved by ADMM. The associated augmented Lagrangian to (\ref{eq:objective_node_full_Z}) reads 
\[ 
\begin{aligned}
\gL(\mU,\mZ; \mY) 
:= &\|\vnu\mZ\|_{p,G}+\tfrac12\|\mM\odot(\mU-\mX)\|^q_{q,G} \\
& + \iprod{\mY}{\wtW\mU-\mZ} + \tfrac{\gamma}{2}\norm{\wtW\mU-\mZ}^2 ,
\end{aligned}
\]
where $\gamma > 0$. 
We consider the inertial ADMM motivated by Alvarez \& Attouch \cite{alvarez2001inertial} to secure an efficient solver, which defines 
\begin{equation} \label{eq:iadmm}
    \begin{aligned} 
      \mZ_{k+1} &= \argmin_{\mZ}~\|\vnu\mZ\|_{p,G}+ \tfrac{\gamma}{2}\norm{\mZ-( 2\mY_{k}-\wtV_{k})/\gamma}^2, \\
      \mV_{k+1} &= \mY_{k}-\gamma \mZ_{k+1}, \\
      \wtV_{k+1} &= \mV_{k+1} + a_k (\mV_{k+1}-\mV_{k}), \\
      \mU_{k+1} &= \argmin_{\mU}~ \tfrac12\|\mM\odot(\mU-\mX)\|^q_{q,G} +\tfrac{\gamma}{2}\norm{\wtW\mU+\wtV_{k+1}/\gamma}^2, \\
      \mY_{k+1} &= \wtV_{k+1}+\gamma \wtW\mU_{k+1}, 
    \end{aligned}
\end{equation}
where $\wtV_{k+1}$ is the inertial term and $a_k$ is the inertial parameter. We refer to \cite{bot2014inertial,boct2015inertial}, and the references therein for the convergence analysis of the above inertial ADMM scheme. 


\begin{table*}[!t]
    \caption{Average performance for node classification over $10$ repetitions. }
    \label{tab:node_classification}
    \begin{center}
    \begin{minipage}{0.95\textwidth}
    \resizebox{\textwidth}{!}{
    \begin{tabular}{lccccccccccc}
    \toprule
    & \multicolumn{8}{c}{\textbf{attribute injection}} & \multicolumn{3}{c}{\textbf{meta attack}} \\ \cmidrule(lr){2-9}\cmidrule(lr){10-12}
    \textbf{Module} & \textbf{Cora} & \textbf{Citeseer} & \textbf{PubMed} & \textbf{Coauthor-CS} & \textbf{Wiki-CS} & \textbf{Wisconsin} & \textbf{Texas} & \textbf{OGB-arxiv}
    & \textbf{Cora} & \textbf{Citeseer} & \textbf{PubMed} \\ 
    \midrule
    clean & $81.26${\scriptsize$\pm0.65$} &
    $71.77${\scriptsize$\pm0.29$} &
    $79.01${\scriptsize$\pm0.44$} &
    $90.19${\scriptsize$\pm0.48$} & 
    $77.62${\scriptsize$\pm0.26$} & 
    $56.47${\scriptsize$\pm5.26$} & $65.14${\scriptsize$\pm1.46$} &
    $71.10${\scriptsize$\pm0.21$}& 
    $81.26${\scriptsize$\pm0.65$} &
    $71.77${\scriptsize$\pm0.29$} &
    $79.01${\scriptsize$\pm0.44$} \\ 
    \midrule
    \textsc{GCN} & $69.06${\scriptsize$\pm0.74$} &
    $57.58${\scriptsize$\pm0.71$} &
    $67.69${\scriptsize$\pm0.40$} &
    $82.41${\scriptsize$\pm0.23$} & 
    $65.44${\scriptsize$\pm0.23$} & 
    $48.24${\scriptsize$\pm3.19$} &
    $58.92${\scriptsize$\pm2.02$} &
    $68.42${\scriptsize$\pm0.15$} &
    $75.07${\scriptsize$\pm0.64$} &
    $55.32${\scriptsize$\pm2.22$} &
    $72.88${\scriptsize$\pm0.30$} \\     
    \textsc{APPNP} & \cellcolor{white!50!green!5!}$68.46${\scriptsize$\pm0.81$} &
    \cellcolor{white!50!red!17!}$60.04${\scriptsize$\pm0.59$} &
    \cellcolor{white!50!red!9!}$68.70${\scriptsize$\pm0.47$} &
    \cellcolor{white!50!green!100!}$71.14${\scriptsize$\pm0.54$} & 
    \cellcolor{white!50!green!70!}$56.53${\scriptsize$\pm0.72$} & 
    \cellcolor{white!50!red!100!}$61.76${\scriptsize$\pm5.21$} &
    \cellcolor{white!50!red!8!}$59.46${\scriptsize$\pm0.43$} &
    \cellcolor{white!50!green!}OOM &
    \cellcolor{white!50!green!25!}$73.49${\scriptsize$\pm0.59$} &
    \cellcolor{white!50!red!2!}$55.67${\scriptsize$\pm0.28$} & 
    \cellcolor{white!50!green!36!}$70.63${\scriptsize$\pm1.07$}\\ 
    \textsc{GNNGuard} & \cellcolor{white!50!green!59!}$61.96${\scriptsize$\pm0.30$} &
    \cellcolor{white!50!green!17!}$54.94${\scriptsize$\pm1.00$} &
    \cellcolor{white!50!red!7!}$68.50${\scriptsize$\pm0.38$} &
    \cellcolor{white!50!green!22!}$80.67${\scriptsize$\pm0.88$} &
    \cellcolor{white!50!red!1!}$65.69${\scriptsize$\pm0.32$} & 
    \cellcolor{white!50!green!16!}$46.86${\scriptsize$\pm1.06$} &
    \cellcolor{white!50!red!4!}$59.19${\scriptsize$\pm0.81$} &
    \cellcolor{white!50!green!78!}$65.75${\scriptsize$\pm0.32$} &
    \cellcolor{white!50!green!50!}$72.02${\scriptsize$\pm0.61$} &
    \cellcolor{white!50!red!14!}$57.64${\scriptsize$\pm1.31$} & 
    \cellcolor{white!50!green!29!}$71.10${\scriptsize$\pm0.32$}\\ 
    \textsc{ElasticGNN} & \cellcolor{white!50!red!71!}$77.74${\scriptsize$\pm0.79$} &
    \cellcolor{white!50!red!50!}$64.61${\scriptsize$\pm0.85$} &
    \cellcolor{white!50!red!31!}$71.23${\scriptsize$\pm0.21$} &
    \cellcolor{white!50!green!33!}$79.91${\scriptsize$\pm1.39$} & 
    \cellcolor{white!50!green!10!}$64.18${\scriptsize$\pm0.53$} & 
    \cellcolor{white!50!red!60!}$53.33${\scriptsize$\pm2.45$} &
    \cellcolor{white!50!red!13!}$59.77${\scriptsize$\pm3.24$} &
    \cellcolor{white!50!green!}$41.34${\scriptsize$\pm0.38$}&
    \cellcolor{white!50!red!68!}\bm{$79.25${\scriptsize$\pm0.50$}} &
    \cellcolor{white!50!red!73!}$67.29${\scriptsize$\pm1.17$} & 
    \cellcolor{white!50!green!15!}$71.95${\scriptsize$\pm0.52$}\\
     \textsc{AirGNN} & \cellcolor{white!50!red!58!}$76.22${\scriptsize$\pm3.75$} &
    \cellcolor{white!50!red!32!}$62.14${\scriptsize$\pm0.82$} &
    \cellcolor{white!50!red!62!}$74.73${\scriptsize$\pm0.43$} &
    \cellcolor{white!50!green!28!}$80.18${\scriptsize$\pm0.31$} & 
    \cellcolor{white!50!red!48!}$71.36${\scriptsize$\pm0.20$} & 
    \cellcolor{white!50!red!100!}$61.56${\scriptsize$\pm0.72$} &
    \cellcolor{white!50!red!8!}$59.46${\scriptsize$\pm1.24$} &
    \cellcolor{white!50!green!100!}$52.32${\scriptsize$\pm0.58$}&
    \cellcolor{white!50!red!62!}$78.94${\scriptsize$\pm0.45$} &
    \cellcolor{white!50!red!62!}$65.58${\scriptsize$\pm0.63$} & 
    \cellcolor{white!50!red!92!}$78.58${\scriptsize$\pm0.71$} \\
    \textsc{MAGnet}{\scriptsize{one}} & \cellcolor{white!50!red!56!}$75.88${\scriptsize$\pm0.42$} &
    \cellcolor{white!50!red!11!}$59.22${\scriptsize$\pm0.34$} &
    \cellcolor{white!50!red!12!}$68.97${\scriptsize$\pm0.21$} &
    \cellcolor{white!50!red!21!}$84.04${\scriptsize$\pm0.56$} &
    \cellcolor{white!50!red!44!}$70.83${\scriptsize$\pm0.29$} & 
    \cellcolor{white!50!red!86!}$55.49${\scriptsize$\pm1.53$} &
    \cellcolor{white!50!red!22!}$60.27${\scriptsize$\pm1.73$} &
    \cellcolor{white!50!green!5!}$68.24${\scriptsize$\pm0.30$} &
    \cellcolor{white!50!red!33!}$77.11${\scriptsize$\pm0.45$} &
    \cellcolor{white!50!red!44!}$62.49${\scriptsize$\pm1.70$} &
    \cellcolor{white!50!red!48!}$75.83${\scriptsize$\pm2.05$} \\ 
    \textsc{MAGnet}{\scriptsize{gae}} & \cellcolor{white!50!red!82!}\bm{$79.07${\scriptsize$\pm0.56$}} &
    \cellcolor{white!50!red!51!}\bm{$64.79${\scriptsize$\pm0.73$}} &
    \cellcolor{white!50!red!68!}\bm{$75.41${\scriptsize$\pm0.35$}} &
    \cellcolor{white!50!red!53!}\bm{$86.50${\scriptsize$\pm0.37$}} & 
    \cellcolor{white!50!red!57!}\bm{$72.40${\scriptsize$\pm0.21$}} & 
    \cellcolor{white!50!red!100!}\bm{$64.31${\scriptsize$\pm2.60$}} &
    \cellcolor{white!50!red!31!}\bm{$60.81${\scriptsize$\pm2.18$}} &
    \cellcolor{white!50!red!8!}\bm{$68.68${\scriptsize$\pm0.03$}} &
    \cellcolor{white!50!red!64!}$79.04${\scriptsize$\pm0.50$} &
    \cellcolor{white!50!red!73!}\bm{$67.40${\scriptsize$\pm0.73$}} &
    \cellcolor{white!50!red!94!}\bm{$78.63${\scriptsize$\pm0.32$}}\\ 
    \midrule
    \textsc{MAGnet}{\scriptsize{true}} & 
    \cellcolor{white!50!red!77!}$78.48${\scriptsize$\pm0.67$} & 
    \cellcolor{white!50!red!72!}$68.55${\scriptsize$\pm0.74$} &
    \cellcolor{white!50!red!70!}$75.63${\scriptsize$\pm0.56$} & 
    \cellcolor{white!50!red!88!}$89.23${\scriptsize$\pm0.40$} &
    \cellcolor{white!50!red!83!}$75.50${\scriptsize$\pm0.20$} & 
    \cellcolor{white!50!red!100!}$65.69${\scriptsize$\pm1.57$} &
    \cellcolor{white!50!red!26!}$60.54${\scriptsize$\pm2.16$} &
    \cellcolor{white!50!red!34!}$69.57${\scriptsize$\pm0.23$} &
    \cellcolor{white!50!red!94!}$80.88${\scriptsize$\pm0.37$} & 
    \cellcolor{white!50!red!74!}$67.46${\scriptsize$\pm0.95$} &
    \cellcolor{white!50!red!100!}$79.16${\scriptsize$\pm0.41$} \\
    \bottomrule\\[-2.5mm]
    \end{tabular}
    }
    \end{minipage}
	\begin{minipage}{0.01\linewidth}
		\begin{annotate}{\includegraphics[width=\linewidth]{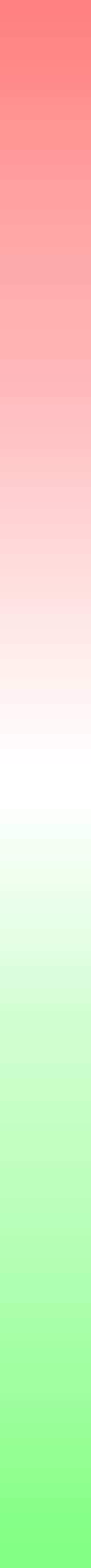}}{1}
    	\note{0.38,1.54}{- $100\%$}
    	\note{0.33,0.77}{- $50\%$}
    	\note{0.28,0.01}{- $0\%$}
    	\note{0.4,-0.76}{- $-50\%$}
    	\note{0.45,-1.515}{- $-100\%$}
    	\end{annotate}
		\label{ }
	\end{minipage}
    \end{center}
\end{table*}

\subsection{Subproblems}
The solutions to the subproblems of $\mZ_{k+1}$ and $\mU_{k+1}$ in (\ref{eq:iadmm}) depend on $p$ and $q$, respectively. Here we enumerate the special cases of $p=\{0,1\}$ and $q=\{1,2\}$. Other choices of $p$, such as $p=\frac12$, are discussed in \cite{xu2012lp} and the references therein.

Different values of $p$ affects the thresholding operator in updating $\mZ_{k+1}$. When $p=1$, $\|\vnu\mZ\|_{p,G}$ becomes the $\ell_1$-norm and its update requires a soft-thresholding, that is,
\[
S_{\alpha}(x) = 
\sign(x) \odot \max\big\{ |x|-\alpha, 0 \big\} . 
\]
When $p=0$, hard-thresholding is applied with $H_{\alpha}(x) = 
\left\{
\begin{aligned}
    x ~&~~ |x| > \alpha \\
    0 ~&~~ |x|\leq \alpha
\end{aligned}
\right.$. 

Compared to $\mZ_{k+1}$, the update of $\mU_{k+1}$ in (\ref{eq:iadmm}) is more complicated as it involves more regularization terms. 
When $q=2$, the objective is a quadratic problem. With $\wtW^{\top}\wtW = \mathrm{Id}$, we directly have
\begin{equation*}  \label{U_q2_1}
    \mU_{k+1} = \frac{\mM\odot\mX-\wtW^{\top}\wtV_{k+1}}{\mM+\gamma}, 
\end{equation*}
which requires an element-wise division. Since the fast approximation of $\wtW$ is implemented by the Chebyshev approximation, it happens when the approximation degree is considerably small that the approximation has a noticeable error, i.e., $\wtW^T\wtW \neq \mathrm{Id}$. Alternatively, the descent type methods such as gradient descent, and conjugate gradient can be applied to inexactly solve the sub-problem with $I$ steps of iteration. At the $i$th ($i\leq I$) iteration,
\begin{equation*} \label{U_q2_2}
    \begin{aligned} 
    \mU^{(i+1)} &= \mU^{(i)} - \alpha \big(\mM\odot(\mU^{(i)}-\mX) + \gamma \wtW^{\top}(\wtW\mU^{(i)} + \wtV_{k+1}/\gamma) \big)
    \end{aligned}
\end{equation*}
where $\alpha$ is the step-size. The solution is then $\mU_{k+1}=\mU^{(I)}$.

In the case of $q=1$, let $\mQ = \mU-\mX$ and consider 
\begin{equation*}
    \tfrac12\|\mM \odot \mQ\|_{1,G} + \tfrac{\gamma}{2}\norm{\mQ+\mX + \wtW^{\top}\wtV_{k+1}/\gamma}^2 . 
\end{equation*}
Let $\mQ_{k+1}$ be the soft-thresholding of $-\mX-\wtW^{\top}\wtV_{k+1}/\gamma$ and we have
\begin{equation} \label{U_q1}
    \mU_{k+1} = \mQ_{k+1}+\mX. 
\end{equation}

\begin{table*}[t]
    \caption{Average performance with different choices on $p$ and $q$ for node classification.}
    \label{tab:ablation_injection}
    \begin{center}
    \resizebox{0.8\textwidth}{!}{
    \begin{tabular}{lcccccccccc}
    \toprule
    \multicolumn{3}{c}{\textbf{Module}} & \multicolumn{5}{c}{\textbf{attribute injection}} & \multicolumn{3}{c}{\textbf{meta attack}} \\  \cmidrule(lr){1-3}\cmidrule(lr){4-8}\cmidrule(lr){9-11}
    mask & reg ($p$) & loss ($q$) & \textbf{Cora} & \textbf{Citeseer} & \textbf{PubMed} & \textbf{Coauthor-CS} & \textbf{Wiki-CS} & \textbf{Cora} & \textbf{Citeseer} & \textbf{PubMed}\\
    \midrule
    N/A & $L_2^{\star}$ & $L_2$ & \cellcolor{gray!25}$69.06${\scriptsize$\pm0.74$} & 
    \cellcolor{gray!25}$57.58${\scriptsize$\pm0.71$} & 
    \cellcolor{gray!25}$67.69${\scriptsize$\pm0.23$} & 
    \cellcolor{gray!25}$82.41${\scriptsize$\pm0.40$} & 
    \cellcolor{gray!25}$65.44${\scriptsize$\pm0.23$} & \cellcolor{gray!25}$75.07${\scriptsize$\pm0.64$} & 
    \cellcolor{gray!25}$55.32${\scriptsize$\pm0.64$} & 
    \cellcolor{gray!25}$72.88${\scriptsize$\pm0.30$} \\
    \midrule
    \multicolumn{1}{l|}{\multirow{4}{*}{\textsc{ONE}}} & \multicolumn{1}{c|}{\multirow{2}{*}{$L_0$}} & $L_1$ & $58.36${\scriptsize$\pm0.89$} & $57.30${\scriptsize$\pm0.74$} & $65.77${\scriptsize$\pm1.03$} & $82.17${\scriptsize$\pm0.41$} & $64.58${\scriptsize$\pm0.21$} & $71.21${\scriptsize$\pm1.12$} & 
    $55.38${\scriptsize$\pm2.12$} & 
    $71.52${\scriptsize$\pm0.43$} \\ 
    \multicolumn{1}{l|}{} & \multicolumn{1}{c|}{} & $L_2$ & $68.74${\scriptsize$\pm0.22$} & $54.07${\scriptsize$\pm1.23$} & $58.52${\scriptsize$\pm1.02$} & $80.63${\scriptsize$\pm0.59$} & $63.63${\scriptsize$\pm0.27$} & $74.46${\scriptsize$\pm0.60$} & 
    $57.75${\scriptsize$\pm2.23$} & 
    $59.76${\scriptsize$\pm1.91$} \\ 
    \multicolumn{1}{l|}{} & \multicolumn{1}{c|}{\multirow{2}{*}{$L_1$}} & $L_1$ & $69.12${\scriptsize$\pm0.57$} & $57.58${\scriptsize$\pm0.71$} & $67.69${\scriptsize$\pm0.40$} & $82.17${\scriptsize$\pm0.41$} & $64.63${\scriptsize$\pm0.18$}  & 
    $75.07${\scriptsize$\pm0.64$} & 
    $55.32${\scriptsize$\pm2.22$} & 
    $72.88${\scriptsize$\pm0.65$} \\ 
    \multicolumn{1}{l|}{} & \multicolumn{1}{c|}{} & $L_2$ & \bm{$75.88${\scriptsize$\pm0.42$}} & \bm{$59.22${\scriptsize$\pm0.34$}} & \bm{$68.97${\scriptsize$\pm0.21$}} & \bm{$84.04${\scriptsize$\pm0.56$}} & \bm{$70.83${\scriptsize$\pm0.29$}} & \bm{$77.11${\scriptsize$\pm0.45$}} & 
    \bm{$62.49${\scriptsize$\pm1.70$}} & 
    \bm{$75.83${\scriptsize$\pm0.35$}} \\ 
    \midrule
    \multicolumn{1}{l|}{\multirow{4}{*}{\textsc{GAE}}} & \multicolumn{1}{c|}{\multirow{2}{*}{$L_0$}} & $L_1$ & $68.42${\scriptsize$\pm1.15$} & $54.38${\scriptsize$\pm0.54$} & $67.74${\scriptsize$\pm0.71$} & $83.95${\scriptsize$\pm0.52$} & $61.96${\scriptsize$\pm0.16$} & $77.42${\scriptsize$\pm1.08$} & 
    $56.13${\scriptsize$\pm0.47$} & 
    $78.42${\scriptsize$\pm0.65$} \\ 
    \multicolumn{1}{l|}{} & \multicolumn{1}{c|}{} & $L_2$ & $66.34${\scriptsize$\pm0.81$} & $56.29${\scriptsize$\pm1.18$} & $59.15${\scriptsize$\pm0.85$} & $83.88${\scriptsize$\pm0.55$} & $64.67${\scriptsize$\pm0.29$} & $77.03${\scriptsize$\pm0.78$} & 
    $55.84${\scriptsize$\pm1.37$} & 
    $70.82${\scriptsize$\pm0.38$} \\ 
    \multicolumn{1}{l|}{} & \multicolumn{1}{c|}{\multirow{2}{*}{$L_1$}} & $L_1$ & $72.76${\scriptsize$\pm0.40$} & $63.60${\scriptsize$\pm0.66$} & \bm{$75.41${\scriptsize$\pm0.35$}} & $86.02${\scriptsize$\pm0.59$} & \bm{$72.40${\scriptsize$\pm0.21$}} & 
    $78.18${\scriptsize$\pm0.56$} & 
    $63.22${\scriptsize$\pm1.56$} & 
    \bm{$78.63${\scriptsize$\pm0.32$}} \\ 
    \multicolumn{1}{l|}{} & \multicolumn{1}{c|}{} & $L_2$ & \bm{$76.81${\scriptsize$\pm0.98$}} & \bm{$64.79${\scriptsize$\pm0.14$}} & $71.13${\scriptsize$\pm0.25$} & \bm{$86.50${\scriptsize$\pm0.37$}} & $60.08${\scriptsize$\pm0.39$} & \bm{$79.04${\scriptsize$\pm0.50$}} & 
    \bm{$67.40${\scriptsize$\pm0.73$}} & 
    $74.47${\scriptsize$\pm0.30$} \\ 
    \midrule
    \multicolumn{1}{l|}{\multirow{4}{*}{\textsc{TRUE}}} & \multicolumn{1}{c|}{\multirow{2}{*}{$L_0$}} & $L_1$ & $77.15${\scriptsize$\pm0.74$} & $68.14${\scriptsize$\pm0.85$} & $74.40${\scriptsize$\pm0.56$} & $88.14${\scriptsize$\pm0.46$} & $72.42${\scriptsize$\pm0.29$} & \bm{$80.88${\scriptsize$\pm0.37$}} &
    \bm{$67.46${\scriptsize$\pm0.95$}} &
    \bm{$79.16${\scriptsize$\pm0.41$}} \\ 
    \multicolumn{1}{l|}{} & \multicolumn{1}{c|}{} & $L_2$ & $76.44${\scriptsize$\pm0.59$} & $65.02${\scriptsize$\pm0.97$} & $68.12${\scriptsize$\pm1.24$} & $87.01${\scriptsize$\pm0.24$} & $71.51${\scriptsize$\pm0.20$}  & $80.57${\scriptsize$\pm0.51$} &
    $67.21${\scriptsize$\pm1.63$} &
    $71.90${\scriptsize$\pm0.41$} \\ 
    \multicolumn{1}{l|}{} & \multicolumn{1}{c|}{\multirow{2}{*}{$L_1$}} & $L_1$ & \bm{$78.48${\scriptsize$\pm0.67$}} & \bm{$68.55${\scriptsize$\pm0.74$}} & \bm{$75.63${\scriptsize$\pm0.56$}} & \bm{$89.23${\scriptsize$\pm0.37$}} & \bm{$75.50${\scriptsize$\pm0.20$}} & $79.99${\scriptsize$\pm0.45$} &
    $65.50${\scriptsize$\pm0.81$} &
    $79.14${\scriptsize$\pm0.32$} \\ 
    \multicolumn{1}{l|}{} & \multicolumn{1}{c|}{} & $L_2$ & $77.57${\scriptsize$\pm0.92$} & $64.29${\scriptsize$\pm0.19$} & $75.19${\scriptsize$\pm0.18$} & $86.72${\scriptsize$\pm0.31$} & $74.44${\scriptsize$\pm0.23$} & $77.16${\scriptsize$\pm0.66$} &
    $67.33${\scriptsize$\pm0.49$} &
    $75.04${\scriptsize$\pm0.32$} \\ 
    \bottomrule\\[-2.5mm]
    \end{tabular}
    }
    \end{center}
\end{table*}

\section{Numerical Experiments}
\label{sec:exp}
This section validates \textsc{MAGnet} by a variety of experiments. We start by comparing our methods to three denoising \ methods on node classification tasks with two types of attribute corruptions. The next two ablation studies digest the influence of regularizers in mask generation and ADMM denoising towards the graph recovery and property prediction performance of the proposed model. To establish an intuitive understanding of the two learning modules, i.e., mask matrix approximation and ADMM optimization, their effects are demonstrated with visualizations.
%
The implementations at \url{https://github.com/bzho3923/MAGnet} are programmed with \texttt{PyTorch-Geometric} (version 2.0.1) and \texttt{PyTorch} (version 1.7.0) and run on NVIDIA$^{\circledR}$ Tesla A100 GPU with $6,912$ CUDA cores and $80$GB HBM2 mounted on an HPC cluster.

\subsection{Experimental Protocol}
\subsubsection{Benchmark Preparation} 
We examine \textsc{MAGnet} on eight publicly available benchmark datasets: \textbf{Cora}, \textbf{Citeseer} and \textbf{PubMed} of the citation networks \cite{yang2016revisiting}; \textbf{Wiki-CS} \cite{mernyei2020wiki} that classifies articles from Wikipedia database; \textbf{Coauthor-CS} \cite{shchur2018pitfalls} that labels the most active field of authors; \textbf{OGB-arxiv} from open graph benchmark \cite{hu2020open} that represents the citation network between all arXiv papers in Computer Science; and \textbf{Wisconsin} and \textbf{Texas} \cite{garcia2016using} that records the web pages from computer science departments of different universities and their mutual links. In particular, the first three datasets are the most classic citation networks that are used for node-level property prediction tasks, \textbf{Wiki-CS} employs dense edge connection, \textbf{Coauthor-CS} elaborates (relatively) high dimension of feature attributes, \textbf{OGB-arxiv} is a large-scale dataset, and the last two web-page datasets are heterophilic graphs.

While the given datasets do not provide ground truth of anomalies, we conduct two types of black-box poisoning methods that have been used in graph anomaly detection and graph defense: 
\begin{itemize}[leftmargin=*]
    \item An \emph{attribute injection} method perturbs attributes through swapping attributes of the most distinct samples in a random subgraph. The noise is added similarly to \cite{ding2019deep}. A certain number of targeted nodes are randomly selected from the input graph. For each selected node $v_i$, we randomly pick another $k$ nodes from the graph and select the node $v_j$ whose attributes deviate the most from node $v_i$ among the $k$ nodes by maximizing the Euclidean distance of node attributes, i.e.,  $\left\|x_{i}-x_{j}\right\|_{2}$. Then, we substitute the attributes $x_{i}$ of node $v_i$ with $x_{j}$. Specifically, we set the size of candidates $k=100$ for small datasets, i.e., \textbf{Cora} and \textbf{Citeseer}, and $k=500$ for relatively larger datasets, i.e., \textbf{PubMed}, \textbf{Coauthor-CS} and \textbf{Wiki-CS}.
    \item Secondly, \emph{meta attack} perturbs node attributes leverages meta-learning in graph adversarial attack to poison node attributes with the meta-gradient of the loss function \cite{zugner2019adversarial} \footnote{implemented by \texttt{DeepRobust} \cite{li2020deeprobust} at \url{https://github.com/DSE-MSU/DeepRobust}}. 
    We use the technique to create local corruptions on the underlying graph by corrupting a small portion of attributes in the graph's feature matrix, where the specific amount of perturbation varies for different datasets to obtain a noticeable attack effect. Instead of unifying the preprocessing procedures (i.e., conduct the same attack step for all the baseline methods), we unify the input graph (i.e., use the same attacked graph input) for baseline methods by fixing the surrogate model to a 2-layer GCN. The justification comes in two-fold. First, we aim to demonstrate each model’s ability to recover a robust representation from given corrupted inputs. For this purpose, it is not required to make baseline-specific attacks to maximize the effect of an adversarial attack. Meanwhile, 
    as we do not assume prior knowledge such as "where were the perturbations from" or "how were the corruptions designed", pre-determining a corrupted graph for all baseline models corresponds to a fair comparison.
\end{itemize}

\subsubsection{Training Setup}
We construct the mask matrix by binary approximation errors from a GCN-based autoencoder constituting $2$ encoding layers following $2$ decoding layers. 
During the optimization, an inertial ADMM is iterated for $15$ times to fastly recover the corrupted graph. The first six datasets follow the standard public split and processing rules in \texttt{PyTorch-Geometric} \cite{fey2019fast} or \texttt{OGB} \cite{hu2020open}. For the two heterophilic datasets, we split the training, validation, and test sets following \cite{pei2020geom}. The reported average test accuracy (or AUROC for \textbf{OGB-arxiv}) is evaluated over $10$ repetitions. 

The models are fine-tuned within the hyper-parameter searching space defined in Table~\ref{tab:searchSpace}, where the first five hyper-parameters are universally applicable to all the models, and the last $\nu_0$ is exclusive to \textsc{MAGnet}. As for model architecture, we construct two convolution layers for \textsc{GCN} (except for \textbf{OGB-arxiv}, where we train a $3$-layer GCN), and three blocks for \textsc{ElasticGNN} and \textsc{APPNP} where each block contains a fully-connected layer and a graph convolution. For \textsc{GNNGuard} and \textsc{MAGnet}, we run the main layers to find the denoised representation $\mU$ and send it to a $2$-layer GCN for node classification. The prediction results are activated by a softmax function for label assignment.

\begin{table}[ht]
    \caption{Hyperparameter searching space.}
    \label{tab:searchSpace}
    \begin{center}
    \begin{tabular}{lr}
    \toprule
    \textbf{Hyperparameters}  & \textbf{Searching Space}  \\ \midrule
    learning rate  & $10^{-3}$, $5\times 10^{-3}$, $10^{-4}$ \\
    weight decay ($L_2$) \; & $10^{-3}$, $5\times 10^{-3}$, $10^{-4}$ \\
    hidden size & $64$, $128$ \\
    dropout ratio & $0.5$ \\
    epochs & $200$, $500$ \\ 
    $\nu_0$ & $10$, $100$, $500$ \\
    $\tau$ & $0.1$ \\
    \bottomrule
    \end{tabular}
    \end{center}
\end{table}

\subsubsection{Baseline Comparison}
A complete investigation on \textsc{MAGnet} is conducted with three different types of mask matrix: all-ones (\textsc{MAGnet}{\scriptsize{one}}), GAE-approximated (\textsc{MAGnet}{\scriptsize{gae}}), and ground truth (\textsc{MAGnet}{\scriptsize{true}}) matrices. The last case advises the upper limit of our model with the `perfectly' approximated mask matrix. We compare our model to three popular graph smoothing models with different design philosophies: \textsc{APPNP} \cite{klicpera2018predict}~\footnote{\url{https://github.com/klicperajo/ppnp}} 
avoids global smoothness with residual connections; \textsc{GNNGuard} \cite{zhang2020gnnguard} ~\footnote{\url{https://github.com/mims-harvard/GNNGuard}} 
modifies the neighbor relevance in message passing to mitigate local corruption; \textsc{ElasticGNN} \cite{liu2021elastic}~\footnote{\url{https://github.com/lxiaorui/ElasticGNN}} and \textsc{AirGNN} \cite{liu2021graph}~\footnote{\url{ https://github.com/lxiaorui/AirGNN}}
pursues local smoothness with a mixture of $\ell_1$ and $\ell_2$ regularizers. We also train a $2$-layer GCN \cite{Kipf2016semi} as the baseline method, which smooths graph signals by minimizing the Dirichlet energy in the spatial domain.

\begin{figure*}[ht]
    \centering
    \begin{minipage}{0.63\textwidth}
    	\includegraphics[width=0.5\linewidth]{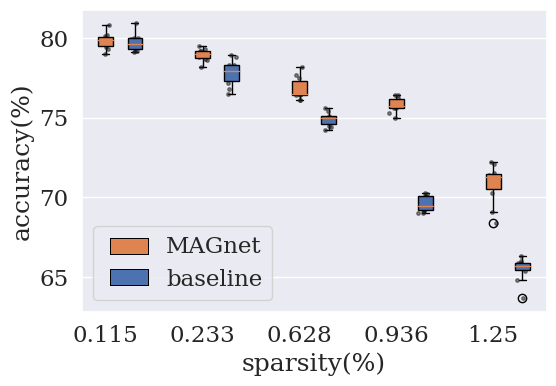}
    	\includegraphics[width=0.5\linewidth]{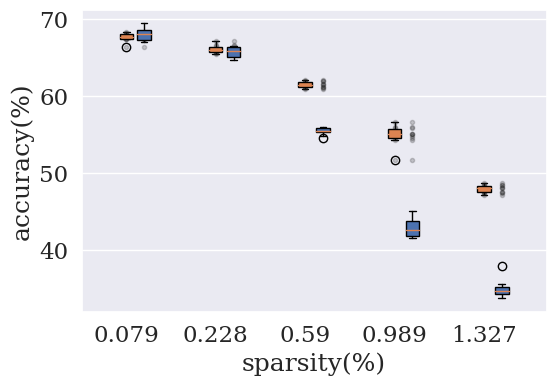}
        \caption{The performance comparison of \textsc{MAGnet} to baseline \textsc{GCN} (corrupted) with different levels of attribute injection to \textbf{Cora} (left) and \textbf{Citeseer} (right).}
        \label{fig:ablation}
    \end{minipage}
    \hfill
    \begin{minipage}{0.35\textwidth}
        \includegraphics[width=\linewidth]{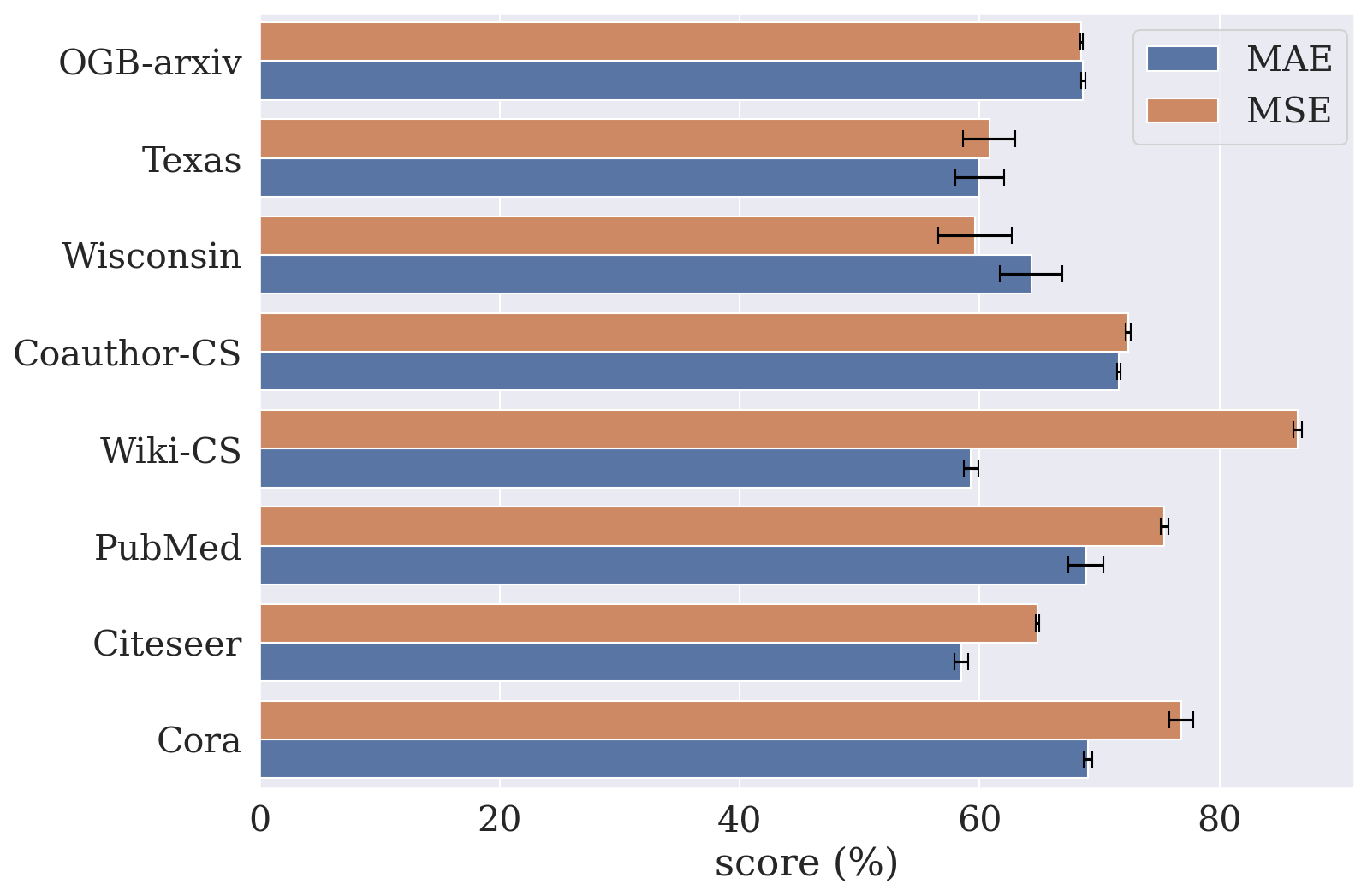}
        \vspace{-7mm}
        \caption{$\ell_1$/$\ell_2$ approximation error in GAE affects the prediction performance.}
        	\label{fig:l1l2}
    \end{minipage}
\end{figure*}

\subsection{Node Classification with Graph Recovery}
Table~\ref{tab:node_classification} compares model performance under two types of local corruptions, i.e., attribute injection and graph meta attack. 
We highlight each score in red or green by their relative improvement or retrogression over GCN. The opacity is determined by the relative percentage change. 

We observe that \textsc{MAGnet}{\scriptsize{gae}} outperforms its competitors and recovers at most $94\%$ test accuracy from the perturbed attributes for homophilic graphs and over $180\%$ for heterophilic graphs. In particular, the results on \textbf{Wisconsin} by \textsc{MAGnet} is even higher than on the non-corrupted clean graph. It implies the potential of our learning scheme to handle heterophilic graphs. We will leave further investigations to the future regarding our denoising model's representation power in handling heterophilic graphs.

On the other hand, the four baseline methods fail to recover a robust embedding from the locally-corrupted input. In some cases, they are fooled to make worse predictions, as they are not able to distinguish outliers from regular patterns. \textsc{APPNP} leverages residual connections to avoid global smoothness by reinforcing local anomalies; \textsc{GNNGuard} makes modifications on the edge connection to indirectly influence node representations; \textsc{ElasticGNN}, although realizing the importance of local smoothness, designs the optimization with low-pass filters and restricts stringent consistency of the new representation even on anomalous positions.

Not only the \textsc{MAGnet} achieves promising performance in graph recovery, but also it requires comparable computational cost. Table~\ref{tab:computationalTime} displays the performance versus running time of the entire program on \textbf{Cora} (with $200$ epochs) and \textbf{OGB-arxiv} (with $500$ epochs). We train GAE with $1,000$ epochs and $10,000$ epochs on the two datasets, respectively. On the small dataset \textbf{Cora}, the running speed of \textsc{GNNGuard} is significantly faster (at $7.35$ seconds) than its counterparts (at around $46$ seconds) at the cost of the lowest accuracy score. Alternatively, \textsc{MAGnet}{\scriptsize{gae}} achieves the best performance with a comparable speed of $45.65$ seconds. On the large-scale dataset \textbf{OGB-arxiv}, the advantage of \textsc{GNNGuard} on the training speed is suppressed by \textsc{ElasticGNN} and \textsc{AirGNN}, but it achieves a higher score than the two baselines. Still, it is $6\%$'s lower than our \textsc{MAGnet}{\scriptsize{gae}} at the cost of a slightly longer training time. As for \textsc{APPNP}, it unfortunately exceeds the memory limit of $100$GB, and thus fails to present any results. In addition to directly making the comparison of our method to baseline models, we also report the training time for \textsc{MAGnet}{\scriptsize{true}}, which is essentially the training cost for denoising the graph with a given mask, i.e., exclude training GAE. Our \textsc{MAGnet} can be potentially more efficient by employing a fast mask approximation tool. Otherwise, a smaller number of training epochs in GAE could also accelerate the overall speed. 

\begin{table}[!thbp]
    \caption{Comparison of the computational cost.}
    \centering
    \resizebox{0.9\linewidth}{!}{
    \begin{tabular}{lcccc}
        \toprule
        & \multicolumn{2}{c}{\textbf{Cora}}  & \multicolumn{2}{c}{\textbf{OGB-arxiv}}  \\ \cmidrule(lr){2-3}\cmidrule(lr){4-5}
        & score (\%) & time (sec) & score (\%) & time (min)\\
        \midrule
        \textsc{APPNP} & $68.46$ & $47.15$ & OOM & OOM\\
        \textsc{GNNGuard} & $61.96$ & $7.35$ & $ 65.75$ & $35.27$ \\
        \textsc{ElasticGNN} & $77.74$ & $45.22$ & $41.34$ & $21.13$ \\
        \textsc{AirGNN} & $76.22$ & $43.92$ & $52.32$ & $11.84$\\
        \textsc{MAGnet}{\scriptsize{gae}} & $79.07$ & $45.65$ & $ 68.68$ & $36.30$ \\
        \textsc{MAGnet}{\scriptsize{true}} & $78.48$ & $33.58$ & $69.57$ & $30.88$ \\
        \bottomrule
    \end{tabular}
    }
    \label{tab:computationalTime}
    \vspace{-7mm}
\end{table}


\subsection{Ablation 1: Reconstruction Error in GAE}
The choices on the regularizers of the mask matrix approximation affect the approximation results. We make comparisons to $\gL=\|\mX-\mX'\|_1$ (MAE loss) and $\gL=\|\mX-\mX'\|_2$ (MSE loss) in GAE. The experiment is conducted on the seven datasets with injection perturbations, with the prediction accuracy reported in Figure~\ref{fig:l1l2}. 

While the empirical performance disagrees with an arbitrary choice of one over another, it supports our analysis in Section~\ref{sec:gae} that MSE loss is generally a more safe choice. More specifically, we observe that $\ell_2$-norm is more likely to outperform when the feature scale is relatively small (e.g., between -1 and 1, which gives a higher punishment to considerably small errors than $\ell_1$'s). As a reference, the attribute values are $\{0,1\}$ (discrete values) for \textbf{Cora}, \textbf{Citeseer}, and \textbf{Coauthor-CS}, [0,1.226] (continuous values) for \textbf{PubMed}, and [-3,3] (continuous values) for \textbf{Wiki-CS}. On the other hand, $\ell_1$-norm tends to perform better in heterophilic graphs, such as \textbf{Wisconsin}. In addition, $\ell_1$-norm is less sensitive to the choice of threshold in (\ref{eq:M}). According to Figure~\ref{fig:tp_fn_mask}, the true positive and false negative rate for the $\ell_1$-based mask matrix does not change drastically among different $\tau$s. In the contrast, tuning the threshold $\tau$ is critical for the $\ell_2$-based scoring function to perform a higher true positive rate and lower false negative rate. When the threshold is properly chosen, its approximation is more reliable than the $\ell_1$-based counterparts.

\subsection{Ablation 2: Regularizers in Graph Recovery}
Table~\ref{tab:ablation_injection} delivers the individual accuracy score for a combination of $p\in\{0,1\}$ and $q\in\{1,2\}$ with all-ones, GAE-oriented, and ground truth mask matrices to digest the preference of $p,q$'s choice in different scenarios for two types of corruptions, respectively. We also include the baseline (GCN) at the top, which is equivalent to optimizing a $p=2$, $q=2$ regularization in the spatial domain.

Overall, $p=1$, $q=2$ is a robust choice when the quality of mask matrix approximation is concerned. The next best choice is $p=1$, $q=1$, which is a frequently-adopted choice in conventional denoising tasks.
When the mask is believed reliable, $p\in\{0,1\}$ and $q=1$ both achieve satisfactory performance. Generally, $p=0$ fits better to concentrated perturbations in a few entities, while $p=1$ fits the case when the corruptions are rather diverse.

Figure~\ref{fig:ablation} reports the effect of corruption level, i.e., the sparsity of the ground truth mask, on the performance with \textbf{Cora} and \textbf{Citeseer}. We visualize the results of \textsc{MAGnet}{\scriptsize{gae}} (orange) with the best-tuned $p,q$. 
The performance gain of \textsc{MAGnet} over GCN on both datasets keeps rising along with an increasing level of corruption.

\begin{figure}[!t]
    \begin{annotate}{\includegraphics[width=0.85\linewidth,trim={0 0 0 8mm},clip]{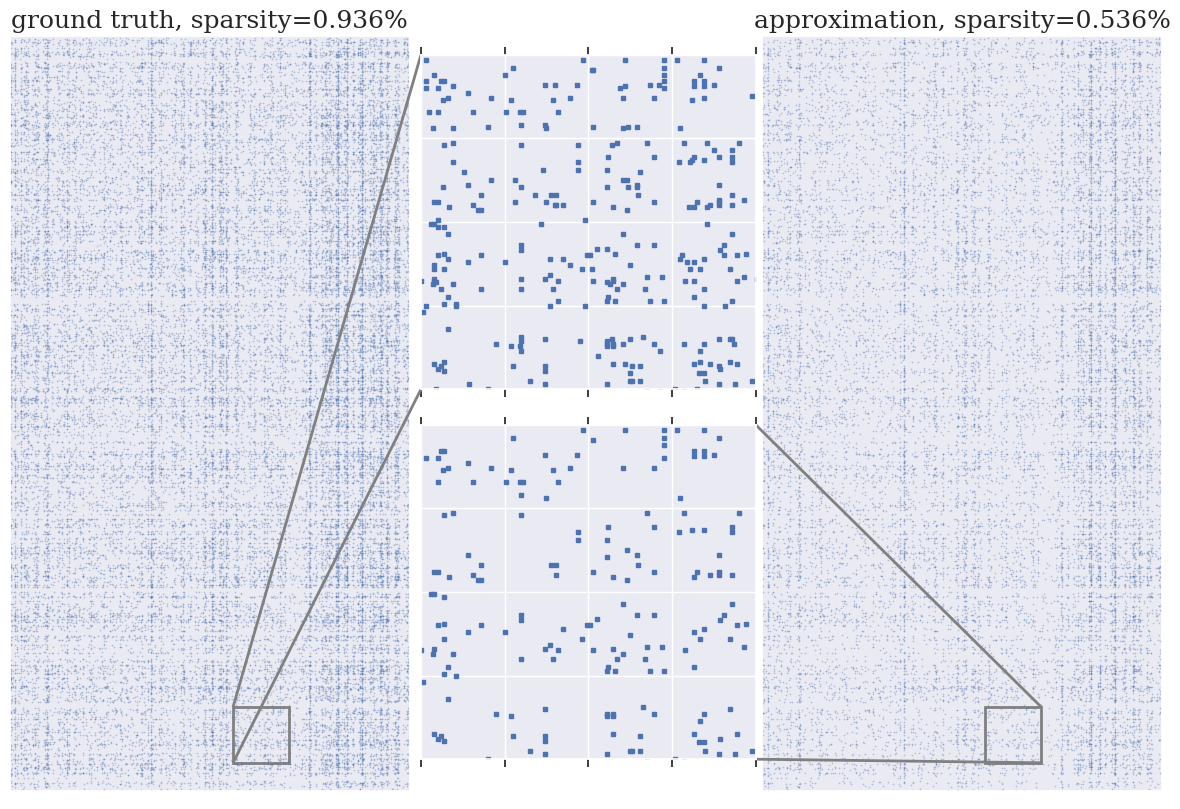}}{1}
    	\note{-2.4,2.65}{sparsity=$0.936\%$}
    	\note{2.4,2.65}{sparsity=$0.536\%$}
    \end{annotate}
    \vspace{-2mm}
	\caption{Distribution of the ground truth mask (left) and conditional GAE-approximated mask (right) at threshold $\tau=0.1$. A $200\times 200$ sub-region is amplified in the middle.}
    \label{fig:spy_cora}
\end{figure}

\subsection{Investigation on Mask Matrix and Local Regularization}
The next experiment digests the effect of the GAE-oriented mask approximation and ADMM optimization with visualizations. 

The quality of GAE is verified by the recall of the mask approximation. Figure~\ref{fig:spy_cora} pictures the conditional mask matrix from model reconstruction error on \textbf{Cora} with attribute injection. 
The sparsity of both matrices can be identified clearly in the amplified subfigures. The approximated mask matrix succeeds in seizing $60\%$ of anomalies, which provides a reliable foundation for subsequent recovering work. 

\begin{figure}[!tbp]
    \begin{annotate}{\includegraphics[width=0.9\linewidth]{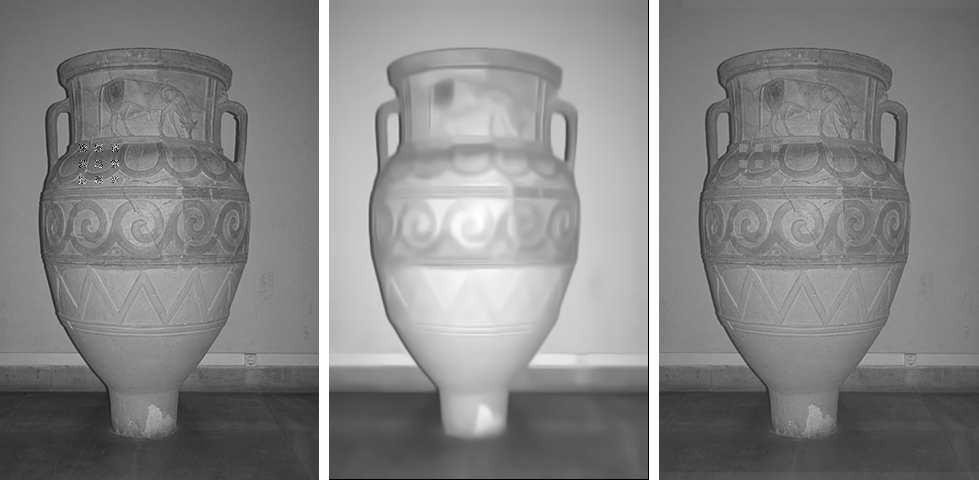}}{1}
        \draw[red,thick] (-3.25,0.78)--++(0.4,0)--++(0,-0.4)--++(-0.4,0)--cycle;
        \note{-2.6,-2.2}{Original image}
        \note{-2.6,-2.5}{Noise in Red Box}
        \note{0,-2.2}{Local PSNR=$17.43$}
        \note{0,-2.5}{Global PSNR=$27.64$}
        \note{2.6,-2.2}{Local PSNR=$24.59$}
        \note{2.6,-2.5}{Global PSNR=$33.81$}
    \end{annotate}
    \caption{Image recovery with local additive white noise at $\sigma=50$. Three images display the noisy raw input (left), and the inpainting result by \textsc{BM3D} (middle) and \textsc{MAGnet} (right).}
    \label{fig:admm}
    \vspace{-4mm}
\end{figure}

Figure~\ref{fig:admm} visually exhibits the local effect of ADMM optimization with an example of graph inpainting, where a raw picture is chosen from the \textbf{BSD68} dataset with $480\times 320$ pixels. It is processed to a graph of $2,400$ nodes and $64$ feature attributes. Each node is transformed by a patch of $8\times 8$ pixels. A lattice-like graph adjacency matrix is prepared with patch neighbors connected with edges. We randomly select $9$ nodes to assign white noise of $\gN(0,1)$ on their attributes. We assume a given mask matrix and focus on the performance of the ADMM optimizer. 
\textsc{MAGnet} restricts major smoothing effects within the masked region. The rest `clean' area maintains a sharp detail. In comparison, the classic denoising model \textsc{BM3D} \cite{dabov2007image} blurs the entire scale of the picture.

\section{Related Work}
\label{sec:review}
The last few years have witnessed a roaring success of graph neural networks, the essential design of which is a graph convolution operation. Depending on whether an adjacency matrix or a graph Laplacian is explicitly employed, a convolution operation falls into two genres of either spectral \cite{bruna2013spectral,defferrard2016convolutional,xu2018graph,zheng2021framelets} or spatial-based \cite{Kipf2016semi,velivckovic2017graph,hamilton2017inductive,monti2017geometric,li2018deeper} methods. 
The widespread message passing scheme \cite{gilmer2017neural} encourages continuous development on adjacency matrix-based feature aggregation rules \cite{hamilton2017inductive,Kipf2016semi,li2018deeper,monti2017geometric,velivckovic2017graph}. These spatial methods pursue global feature smoothness by minimizing Dirichlet energy, which establishes a low-pass filter that denoises under $\ell_2$ penalty \cite{dong2021adagnn,nt2019revisiting,zhu2021interpreting}.
When multiple convolution layers are stacked in a neural network, it is inevitable to lose small-energy patterns and learn an over-smoothing representation. In contrast, residual connections \cite{chen2020simple,klicpera2018predict,rong2019dropedge,xu2018representation} maintain feature proximity and concurrently keep odd examples, i.e., anomalies, that potentially disturb pattern characterization. 

The harmful graph anomalies are getting noticed in the literature. Robust graph optimization refines the learning manner of a classifier to provide promising predictions against potential threats \cite{dai2018adversarial,xu2020adversarial,zugner2019adversarial}. They are essentially designed to defend against graph adversarial attacks, which usually influence model performance by corrupting node connections. A robust node embedding is indirectly approximated by a refined adjacency matrix \cite{chen2020iterative,entezari2020all,jin2020graph,zhang2020gnnguard}. 
Graph anomaly detection \cite{ding2019deep,ma2021comprehensive,peng2020deep,zhu2020anomaly} identifies rare problematic nodes, but the related methods merely recover robust representation of anomalies. On top of that, the justification is based on node entities rather than feature values. 
A few existing works modify the loss function to force local smoothness of the feature space \cite{chen2021graph,liu2021graph,liu2021elastic}, but they limit the regularization in the spatial domain, and the penalties are restricted to $\ell_1$ or $\ell_2$.

\section{Conclusion}
\label{sec:conclusion}
This research develops a GNN-based framework to recover robust graph data representations from locally corrupted node attributes. 
We propose a multi-scale sparse regularizer to optimize the hidden representation that guarantees conditional closeness to the disrupted input. The outliers in graph node attributes are explicitly positioned by a learnable mask matrix, which is approximated by an unsupervised graph autoencoder that requires no prior knowledge of the distribution of anomalies. 
We define the optimization problem by adaptive $l_p$ and $l_q$ errors, where the tunable factors $p,q$ stimulate the maximum ability of the regularizer, and the corresponding optimization can be solved by an efficient inertial ADMM. Incorporating sparsity in multi-level framelet coefficients not only removes global noises and regularizes local anomalous attributes at the same time but also preserves local patterns.  
Our model achieves competitive performance in recovering a robust graph embedding from local corruptions, whereas graph smoothing and defense baseline methods fail to provide a decent solution.

\begin{acks}
This work was supported by the National Natural Science Foundation of China (No. 62172370) and Shanghai Artificial Intelligence Laboratory (No. P22KN00524).
\end{acks}

\bibliographystyle{ACM-Reference-Format}
\bibliography{main.bbl}

\appendix
\section{Graph Framelet Transform}
\label{sec:app:fgt}
This section briefs the fast computation of undecimated framelet transform on graphs. The initial idea was proposed by Dong \cite{dong2017sparse}, which is then developed into a graph convolution \cite{zheng2021framelets}. Here we give a brief introduction to the wavelet frame (framelet) system for a graph signal, using the fast approximation on the framelet decomposition and reconstruction operators by Chebyshev approximation, which is the key to an efficient graph convolution. 

\subsection{Framelet System}
A framelet is defined by two key elements: a \emph{filter bank} $\boldsymbol{\eta}:=\{a;b^{(1)},\dots,b^{(K)}\}$ and a set of \emph{scaling functions} $\Psi=\{\alpha;\beta^{(1)},\dots,\beta^{(K)}\}$. We name $a$ the low-pass filter and $b^{(k)}$ the $k$th high-pass filter with $k=1,\dots,K$. The two sets of filters respectively extract the approximated and detailed information of the input graph signal in a transformed domain, i.e., the framelet domain. 

There are different choices of filter masks, which results in different tight framelet systems (See \cite{dong2017sparse} for some examples). In general, the selected filter bank and its associated scaling function satisfy the relationship that
\begin{equation} \label{eq:relations}
    \widehat{\alpha}(2\xi)=\widehat{a}(\xi) \widehat{\alpha}(\xi),\quad
    \widehat{\beta^{(k)}}(2 \xi)=\widehat{b^{k}}(\xi) \widehat{\alpha}(\xi),
\end{equation}
for $k=1, \ldots, K$ and $\xi \in \mathbb{R}$.
We implement the \emph{Haar-type} filter with one high pass, i.e., $K=1$. For $x\in\R$, it defines
\begin{equation*}
    \widehat{\alpha}(x) = \cos(x/2)\mbox{~~and~~} \widehat{\beta^{(1)}}(x) = \sin(x/2).
\end{equation*}

With the defined filter bank and the scaling functions, the undecimated framelet basis can be defined for a graph signal with the eigenpairs $\{(\lambda_\ell,\vu_\ell)\}_{\ell=1}^{n}$ of the graph Laplacian $\gL$. At level $l=1,\dots,J$, we define the undecimated framelets for node $p$ by
\begin{equation}
  \begin{array}{l}
  \boldsymbol{\varphi}_{l,p}(v):=\sum_{\ell=1}^{n} \widehat{\alpha}\left(\frac{\lambda_{\ell}}{2^{l}}\right) \overline{\vu_{\ell}(p)} \vu_{\ell}(v), \\[2mm]
  \boldsymbol{\psi}_{l,p}^{(k)}(v):=\sum_{\ell=1}^{n} \widehat{\beta^{(k)}}\left(\frac{\lambda_{\ell}}{2^{l}}\right) \overline{\vu_{\ell}(p)} \vu_{\ell}(v), \quad k=1, \ldots, K.
  \end{array}
\end{equation}
We name $\boldsymbol{\varphi}_{l,p}(v),\boldsymbol{\psi}_{l, p}^{(k)}(v)$ with $v\in\gV$ the low-pass and the $k$th high-pass framelet basis, respectively. These bases are called the \emph{undecimated tight framelets} on $\gG$. They define an \emph{undecimated framelet system} $\operatorname{UFS}_{J_1}^{J}(\Psi, \boldsymbol{\eta})$ ($J > J_1$) for $l_2(\gG)$ from $J_1$, which reads
\begin{equation}
\begin{aligned}
  \operatorname{UFS}_{J_1}^{J}(\Psi, \boldsymbol{\eta}) &:=\operatorname{UFS}_{J_1}^{J}(\Psi, \boldsymbol{\eta}; \gG) \\
  &:=\left\{\varphi_{J_1, p}: p \in \gV \right\} \cup\{\psi_{l,p}^{(k)}: p \in \gV, l=J_1,\dots,J\}_{k=1}^{K}. \notag
\end{aligned}
\end{equation}

\subsection{Framelet Decomposition}
We now introduce the \emph{framelet decomposition operator} of the defined tight framelet system, i.e., $\boldsymbol{\gW}$ that we used in \eqref{eq:objective_intro} to construct the spectral coefficients in the framelet domain.

The framelet decomposition operator $\boldsymbol{\gW}_{k,l}$ contains a set of orthonormal bases at $(k,l) \in \{(0, J)\}\cup\{(1,1), \dots,(1, J), 
\dots,(K, J)\}$. It transforms a given graph signal $\mX$ to a set of multi-scale and multi-level \emph{framelet coefficients}, i.e., the spectral representation of the graph signal $\mX$ in the transformed domain. In particular, $\boldsymbol{\gW}_{0,J}$ contains $\boldsymbol{\varphi}_{J,p}, p\in\gV$ that constructs the low-pass framelet coefficients $\boldsymbol{\gW}_{0, J}\mX$ that preserve approximate information in $\mX$. They are the smooth representation that reflects the global trend of $\mX$. Meanwhile, the high-pass coefficients $\boldsymbol{\gW}_{k,l}\mX$ with $\boldsymbol{\gW}_{k,l}=\{\boldsymbol{\psi}_{l,p}^{(k)}, p\in\gV\}$ records detailed information at scale $k$ and level $l$. They reflect local patterns or noises of the signal. A larger scale $k$ contains more localized information with smaller energy.

The framelet coefficients can be directly projected by $\langle\boldsymbol{\varphi}_{l,p},\mX\rangle$ and $\langle\boldsymbol{\psi}_{l,p}^{(k)},\mX\rangle$ for node $p$ at scale level $l$. For instance, we take eigendecomposition on a graph Laplacian $\gL$ and obtain $\mU=[\vu_1,\dots,\vu_n]\in\R^{n\times n}$ be the eigenvectors and $\Lambda=\operatorname{diag}(\lambda_1,\dots,\lambda_n)$ be the eigenvalues. According to the definition above, the respective filtered diagonal matrices with low-pass and high-pass filters are
\begin{equation*}
\begin{aligned}
    \widehat{\alpha}\left(\frac{\Lambda}{2}\right)&=\operatorname{diag}\left(\widehat{\alpha}\left(\frac{\lambda_1}{2}\right),\dots,\widehat{\alpha}\left(\frac{\lambda_n}{2}\right)\right), \\
    \widehat{\beta^{(k)}}\left(\frac{\Lambda}{2^l}\right)&=\operatorname{diag}\left(\widehat{\beta^{(k)}}\left(\frac{\lambda_1}{2^l}\right), \dots, \widehat{\beta^{(k)}}\left(\frac{\lambda_n}{2^l}\right)\right).
\end{aligned}
\end{equation*}

The associated framelet coefficients at the low pass and the $k$th high pass are
\begin{equation} \label{eq:ft_initial}
\begin{aligned}
    &\boldsymbol{\gW}_{0,J}\mX=\mU \widehat{\alpha}\left(\frac{\Lambda}{2}\right) \mU^{\top} \mX , \\
    &\boldsymbol{\gW}_{k,l}\mX=\mU \widehat{\beta^{(k)}}\left(\frac{\Lambda}{2^{l+1}}\right) \mU^{\top} \mX 
    \quad \forall l=0,\dots,J.
\end{aligned}
\end{equation}

\subsection{Fast Tight Framelet Transform}
To formulate an efficient framelet decomposition, two strategies are considered, including a recursive formulation on filter matrices, and Chebyshev-approximated eigenvectors.

With a filter bank that satisfies (\ref{eq:relations}), the above decomposition can be implemented recursively by
\begin{equation*}
\begin{aligned}
    \boldsymbol{\gW}_{k,1}\mX=\mU \widehat{\beta^{(k)}}\left(2^{-R}\Lambda\right) \mU^{\top} \mX 
\end{aligned}
\end{equation*}
for the first level ($l=1$), and 
\begin{equation*}
\begin{aligned}
    \boldsymbol{\gW}_{k,l}\mX&=\mU \widehat{\beta^{(k)}}\left(2^{R+l-1}\Lambda\right)\widehat{\alpha}\left(2^{R+l-2}\Lambda\right) \dots \widehat{\alpha}\left(2^{-R}\Lambda\right)
    \mU^{\top} \mX \\
    &\forall l=2,\dots,J,
\end{aligned}
\end{equation*}
where the real-value dilation scale $R$ satisfies $\lambda_{\max} \leq 2^R\pi$.

Furthermore, we employ an $m$-order Chebyshev polynomials approximation for efficient framelet decomposition. It avoids eigendecomposition on graph Laplacian, which can be considerably slow on a large graph.

Denote the $m$-order approximation of $\alpha$ and $\{\beta^{(k)}\}_{k=1}^{K}$ by $\gT_0$ and $\{\gT_{k}\}_{k=1}^{K}$, respectively. The framelet decomposition operator $\boldsymbol{\gW}_{r,j}$ is approximated by
\begin{equation*}
\boldsymbol{\gW}_{k,l}=
    \begin{cases}
    \gT_0\left(2^{-R} \gL\right), & l=1, \\[1mm]
    \gT_r\left(2^{R+l-1} \gL\right) \gT_0\left(2^{R+l-2} \gL\right) \dots \gT_0\left(2^{-R} \gL\right), & l=2,\dots,J.
    \end{cases}
\end{equation*}

We apply the fast-approximated $\boldsymbol{\gW}$ in the $l_p$ penalty term of the objective function \eqref{eq:objective_intro} formulated in Section~\ref{sec:formulation}.

\section{Inertial ADMM algorithm}
\label{sec:app:admm}
This section provides essential details for the inertial ADMM to understand the update rules defined in Section~\ref{sec:admm} of the main text.

\subsection{Inertial ADMM}
We denote $\mZ = \wtW\mU$ and rewrite \eqref{eq:objective_intro} as
\begin{equation*}
    \min_{\mU,\mZ} ~\|\vnu\mZ\|_{p,G}+\tfrac12\|\mM\odot(\mU-\mX)\|^q_{q,G}, ~~\mathrm{such~that} ~ \mZ = \wtW\mU.
\end{equation*}
This forms a standard formulation of problems that can be solved by Alternating Direction Method of Multipliers (ADMM \cite{gabay1976dual}). The associated augmented Lagrangian reads 
\[ 
\begin{aligned}
\gL(\mU,\mZ; \mY) 
:= &\|\vnu\mZ\|_{p,G}+\tfrac12\|\mM\odot(\mU-\mX)\|^q_{q,G} \\
&+ \iprod{\mY}{\wtW\mU-\mZ} + \tfrac{\gamma}{2}\norm{\wtW\mU-\mZ}^2.
\end{aligned}
\]
To find a saddle-point of $\gL(\mU,\mZ; \mY) $, ADMM applies the following iteration
\begin{align*}
    \mZ_{k+1} =& \argmin_{\mZ}~ \|\vnu\mZ\|_{p,G} + \tfrac{\gamma}{2}\norm{\wtW\mU_{k}-\mZ}^2 +\iprod{\mY_{k}}{\wtW\mU_{k}-\mZ}, \\
    \mU_{k+1} =& \argmin_{\mU}~ \tfrac12\|\mM\odot(\mU-\mX)\|^q_{q,G}\\
    &+\tfrac{\gamma}{2}\norm{\wtW\mU-\mZ_{k+1}}^2+ \iprod{\mY_{k}}{\wtW\mU-\mZ_{k+1}},\\
    \mY_{k+1} =& \mY_{k} + \gamma\big(\wtW\mU_{k+1}-\mZ_{k+1}).
\end{align*}

The above iteration can be equivalently written as
\begin{align*}
    \mZ_{k+1} &= \argmin_{\mZ}~ \|\vnu\mZ\|_{p,G} + \tfrac{\gamma}{2}\norm{\wtW\mU_{k}-\mZ + \mY_{k}/\gamma}^2, \\
    \mU_{k+1} &= \argmin_{\mU}~ \tfrac12\|\mM\odot(\mU-\mX)\|^q_{q,G} + \tfrac{\gamma}{2}\norm{\wtW\mU-\mZ_{k+1} + \mY_{k}/\gamma}^2, \\
    \mY_{k+1} &= \mY_{k} + \gamma\big(\wtW\mU_{k+1}-\mZ_{k+1}).
\end{align*}
If we further define 
\[
\mV_{k+1} = \mY_{k} - \gamma \mZ_{k+1},
\]
the above iteration can be reformulated as
\begin{align*}
    \mZ_{k+1} &= \argmin_{\mZ}~ \|\vnu\mZ\|_{p,G} + \tfrac{\gamma}{2}\norm{\mZ - \wtW\mU_{k}-\mY_{k}/\gamma}^2  , \\
    &= \argmin_{\mZ}~ \|\vnu\mZ\|_{p,G} + \tfrac{\gamma}{2}\norm{\mZ - ( 2\mY_{k} - \mV_{k})/\gamma}^2  , \\
    \mV_{k+1} &= \mY_{k} - \gamma \mZ_{k+1} , \\
    \mU_{k+1} &= \argmin_{\mU}~ \tfrac12\|\mM\odot(\mU-\mX)\|^q_{q,G} + \tfrac{\gamma}{2}\norm{\wtW\mU + \mV_{k+1}/\gamma}^2 , \\
    \mY_{k+1} &= \mY_{k} + \gamma\big(\wtW\mU_{k+1}-\mZ_{k+1}) \\
    &= \mV_{k+1} + \gamma \wtW\mU_{k+1} .
\end{align*}

In this paper, we consider the inertial ADMM motivated by \cite{alvarez2001inertial}, whose iteration is provided below: 
\begin{align*}
\mZ_{k+1} &= \argmin_{\mZ}~ \|\vnu\mZ\|_{p,G} + \tfrac{\gamma}{2}\norm{\mZ - ( 2\mY_{k} - \wtV_{k})/\gamma}^2  , \\
\mV_{k+1} &= \mY_{k} - \gamma \mZ_{k+1} , \\
\wtV_{k+1} &= \mV_{k+1} + a_k (\mV_{k+1}-\mV_{k})   ,   \\
\mU_{k+1} &= \argmin_{\mU}~ \tfrac12\|\mM\odot(\mU-\mX)\|^q_{q,G} + \tfrac{\gamma}{2}\norm{\wtW\mU + \wtV_{k+1}/\gamma}^2 , \\
\mY_{k+1} &= \mY_{k} + \gamma\big(\wtW\mU_{k+1}-\mZ_{k+1}) \\
    &= \wtV_{k+1} + \gamma \wtW\mU_{k+1} .
\end{align*}
In general, we have $a_k \in [0, 1]$. When the problem is convex, the convergence can be guaranteed by choosing $a_k \in [0, 1/3]$ \cite{bot2014inertial,boct2015inertial}. 

\subsection{Solution to Subproblem of $q=1$}
When $q=1$, let $\mQ = \mU-\mX$, and consider 
\[
\min_{\mY}~ \tfrac12\|\mM\odot \mY\|_{1,G} + \tfrac{\gamma}{2}\norm{\wtW\mY+\wtW\mX + \wtV_{k+1}/\gamma}^2 .
\]
The optimality condition yields
\[
\begin{aligned}
& \frac12 \mM \odot \partial \|\mM\odot \mQ\|_{1,G} + \gamma \wtW^T (\wtW\mQ+\wtW\mX + \wtV_{k+1}/\gamma) \\
\Longleftrightarrow &~ \frac12 \mM \odot \partial \|\mM\odot \mQ\|_{1,G} + \gamma  (\mQ+\mX + \wtW^T\wtV_{k+1}/\gamma) \\
\Longleftrightarrow &~ \tfrac12\|\mM \odot \mQ\|_{1,G} + \tfrac{\gamma}{2}\norm{\mQ+\mX + \wtW^T\wtV_{k+1}/\gamma}^2 . 
\end{aligned}
\]
From above we have that $\mQ_{k+1}$ is the soft-thresholding of $-\mX - \wtW^T\wtV_{k+1}/\gamma$ and 
\[
\mU_{k+1} = \mQ_{k+1} + \mX. 
\]

\section{Supplementary Material}
\label{sec:app:sm}
We prepare additional experimental results, including data description, details for percentage improvements we colored in Table~\ref{tab:node_classification}, as well as additional visualizations for the mask approximation  
at \url{https://github.com/bzho3923/MAGnet/blob/main/sm.pdf}.

\begin{table*}[!htbp]
\caption{Summary of the datasets for node classification tasks.}
\label{tab:stats:node_classification}
\begin{center}
\resizebox{0.9\textwidth}{!}{
    \begin{tabular}{lcccccccc}
    \toprule
    & \textbf{Cora} & \textbf{Citeseer} & \textbf{PubMed} & \textbf{Wiki-CS} & \textbf{Coauthor-CS} & \textbf{Wisconsin} & \textbf{Texas} & \textbf{OGB-arxiv} \\
    \midrule
    \# Nodes & $2,708$ & $3,327$ & $19,717$ & $11,701$ & $18,333$ & $251$ & $183$ & $169,343$\\
    \# Edges & $5,429$ & $4,732$ & $44,338$ & $216,123$ & $100,227$ & $499$ & $309$ & $1,166,243$ \\
    \# Features & $1,433$ & $3,703$ & $500$ & $300$ & $6,805$ & $1,7033$ & $1,703$ & $128$ \\
    \# Classes & $7$ & $6$ & $3$ & $10$ & $15$ & $5$ & $5$ & $40$ \\
    \# Training Nodes & $140$ & $120$ & $60$ & $580$ & $300$  & $120$ & $87$ & $ 90,941$ \\
    \# Validation Nodes & $500$ & $500$ & $500$ & $1769$ & $200$ & $80$ & $59$ & $29,799$ \\
    \# Test Nodes & $1,000$ & $1,000$ & $1,000$ & $5847$ & $1000$ & $51$ & $37$ & $48,603$ \\
    Label Rate & $0.052$ & $0.036$ & $0.003$ & $0.050$ & $0.016$ & $0.478$ & $0.475$ & $0.537$ \\
    Feature Scale & $\{0,1\}$ & $\{0,1\}$ & $[0, 1.263]$ & $[-3,3]$ & $\{0,1\}$ & $\{0,1\}$ & $\{0,1\}$ & $[-1.389,1.639]$\\ 
    \bottomrule
    \end{tabular}
}
\end{center}
\end{table*}

\begin{table*}[!ht]
    \caption{Improvement percentage of average performance for node classification with \textbf{injection}.}
    \label{tab:node_injection_perc}
    \begin{center}
    \resizebox{\textwidth}{!}{
    \begin{tabular}{lrrrrrrrrrrrrrrrr}
    \toprule
    & \multicolumn{2}{c}{\textbf{Cora}}  & \multicolumn{2}{c}{\textbf{Citeseer}} & \multicolumn{2}{c}{\textbf{PubMed}} & \multicolumn{2}{c}{\textbf{Coauthor-CS}} & \multicolumn{2}{c}{\textbf{Wiki-CS}} & \multicolumn{2}{c}{\textbf{Wisconsin}} & \multicolumn{2}{c}{\textbf{Texas}} & \multicolumn{2}{c}{\textbf{OGB-arxiv}}\\ \cmidrule(lr){2-3}\cmidrule(lr){4-5}\cmidrule(lr){6-7}\cmidrule(lr){8-9}\cmidrule(lr){10-11}\cmidrule(lr){12-13}\cmidrule(lr){14-15}\cmidrule(lr){16-17}
    \textbf{Module} & absolute & relative & absolute & relative & absolute & relative & absolute & relative & absolute & relative & absolute & relative & absolute & relative & absolute & relative \\
    \midrule
    \textsc{APPNP} & $-0.95\%$ & $-5.44\%$ & $4.27\%$ & $17.34\%$ & $1.49\%$ & $8.92\%$ & $-13.68\%$ & $-148.29\%$ & $-13.62\%$ & $-73.15\%$ & $28.03\%$ & $159.06\%$ & $0.92\%$ & $8.68\%$ & - & -  \\
    \textsc{GNNGuard} & $-10.36\%$ & $-58.98\%$ & $-4.58\%$ & $-17.38\%$ & $-1.20\%$ & $7.16\%$ & $-2.11\%$ & $-22.37\%$ & $0.00\%$ & $0.00\%$ & $-2.86\%$ & $-16.24\%$ & $0.46\%$ & $4.34\%$ & $-3.90\%$  & $-78.07\%$\\ 
    \textsc{ElasticGNN} & $12.47\%$ & $71.00\%$ & $12.21\%$ & $49.54\%$ & $5.23\%$ & $31.27\%$ & $-3.03\%$ & $-32.89\%$ & $-1.93\%$ & $-10.34\%$ & $10.55\%$ & $59.88\%$ & $1.44\%$ & $13.67\%$  & $-39.58\%$ & $-791.81\%$\\ 
    \textsc{AirGNN} & $10.37\%$ & $58.69\%$	& $7.92\%$ & $32.14\%$ & $10.40\%$ & $62.19\%$ & $-2.71\%$ &	$-28.66\%$ & $9.05\%$ & $48.60\%$ &	$27.61\%$ & $161.85\%$ & $0.92\%$ & $8.68\%$ & $-23.53\%$ & $-600.75\%$\\
    \textsc{MAGnet}{\scriptsize{-one}} & $9.78\%$ & $55.68\%$ & $2.85\%$ & $11.31\%$ & $1.89\%$ & $10.80\%$ & $1.98\%$ & $20.95\%$ & $8.24\%$ & $44.25\%$ & $15.03\%$ & $85.29\%$ & $2.29\%$ & $21.70\%$ & $-0.26\%$  & $-5.26\%$\\
    \textsc{MAGnet}{\scriptsize{-gae}} & $14.49\%$ & $82.05\%$ & $12.52\%$ & $50.81\%$ & $11.40\%$ & $68.20\%$ & $4.96\%$ & $52.57\%$ & $10.64\%$ & $57.14\%$ & $33.31\%$ & $189.06\%$ & $3.21\%$ & $30.39\%$  & $0.38\%$ & $7.60\%$\\
    \midrule
    \textsc{MAGnet}{\scriptsize{-true}} & $13.50\%$ & $76.85\%$ & $19.05\%$ & $72.22\%$ & $11.73\%$ & $70.14\%$ & $8.28\%$ & $87.66\%$ & $15.37\%$ & $82.59\%$ & $36.17\%$ & $205.29\%$ & $2.75\%$ & $26.05\%$ & $1.68\%$ & $33.63\%$ \\
    \bottomrule\\[-2.5mm]
    \end{tabular}
    }
    \end{center}
\end{table*}

\begin{table}[ht]
    \caption{Improvement percentage of average performance for node classification with \textbf{meta-attack}.}
    \label{tab:node_mettack_perc}
    \begin{center}
    \resizebox{0.95\linewidth}{!}{
    \begin{tabular}{lrrrrrr}
    \toprule
    & \multicolumn{2}{c}{\textbf{Cora}}  & \multicolumn{2}{c}{\textbf{Citeseer}} & \multicolumn{2}{c}{\textbf{PubMed}} \\ \cmidrule(lr){2-3}\cmidrule(lr){4-5}\cmidrule(lr){6-7}
    \textbf{Module} & absolute & relative & absolute & relative & absolute & relative \\
    \midrule
    \textsc{APPNP} & $-2.10\%$ & $-25.53\%$ & $0.63\%$ & $2.13\%$ & $-3.09\%$ & $-36.70\%$ \\ 
    \textsc{GNNGuard} & $-4.06\%$ & $-49.27\%$ & $4.19\%$ & $14.10\%$ & $-2.44\%$ & $-29.04\%$ \\ 
    \textsc{ElasticGNN} & $5.57\%$  & $67.53\%$ & $21.64\%$ & $72.77\%$ & $-1.28\%$ & $-15.17\%$ \\ 
    \textsc{AirGNN} & $5.16\%$ & $62.52\%$ & $18.55\%$ & $62.37\%$ & $7.82\%$ & $92.99\%$ \\
    \textsc{MAGnet}{\scriptsize{one}} (ours) & $2.72\%$  & $32.96\%$ & $12.96\%$ & $43.59\%$  & $4.05\%$ & $48.12\%$ \\
    \textsc{MAGnet}{\scriptsize{gae}} (ours) & $5.29\%$  & $64.14\%$ & $21.84\%$ & $73.43\%$ & $7.89\%$ & $93.80\%$\\
    \midrule
    \textsc{MAGnet}{\scriptsize{true}} & $7.74\%$ & $93.86\%$ & $21.95\%$ & $73.80\%$ & $8.62\%$ & $102.45\%$\\
    \bottomrule\\[-2.5mm]
    \end{tabular}
    }
    \end{center}
\end{table}

\newpage
\section{Dataset Descriptions}
Table~\ref{tab:stats:node_classification} documents key descriptive statistics of the eight datasets for the node classification tasks. The first five datasets are downloaded from \texttt{PyTorch-Gometric} \cite{fey2019fast} \footnote{\url{https://pytorch-geometric.readthedocs.io/en/latest/modules/datasets.html}}, the two heterophilic graphs are from \texttt{Geom-GCN} \cite{pei2020geom} \footnote{\url{https://github.com/graphdml-uiuc-jlu/geom-gcn}}, and \textbf{OGB-arxiv} is from open graph benchmarks \cite{hu2020open} \footnote{\url{https://ogb.stanford.edu/docs/nodeprop/\#ogbn-arxiv}}. All the preprocessing, data split, and evaluation are guided by the source codes or standard routines.

\section{Percentage Performance Improvement in Node Classification}
\label{sec:app:perc}
In the main paper, we use the relative scores of improvements to color the performance in Table~\ref{tab:node_classification}, which is calculated by 
\begin{equation*}
    \text{Relative Score}=\frac{S-S_{\text{corrupted}}}{S_{\text{clean}}-S_{\text{corrupted}}},
\end{equation*}
where $S$ denotes the current score of the performed model, and $S_{\text{clean}}$ and $S_{\text{corrupted}}$ are the accuracy score of \textsc{GCN} on the clean dataset and corrupted dataset, respectively. We also report the absolute score of improvement for a clear comparison, which is defined by 
\begin{equation*}
    \text{Absolute Score}=\frac{S-S_{\text{corrupted}}}{S_{\text{corrupted}}}.
\end{equation*}
Both scores are reported in Table~\ref{tab:node_injection_perc} below. In Table~\ref{tab:node_classification}, we color the model performance by red for positive relative scores, and by green for negative relative scores. If the absolute value of the relative score is large, the color would be darker.



\section{GAE visualization}
\label{sec:app:gae}
\begin{figure}[t]
    \centering
    \begin{annotate}{\includegraphics[width=\linewidth,trim={0 0 0 9mm},clip]{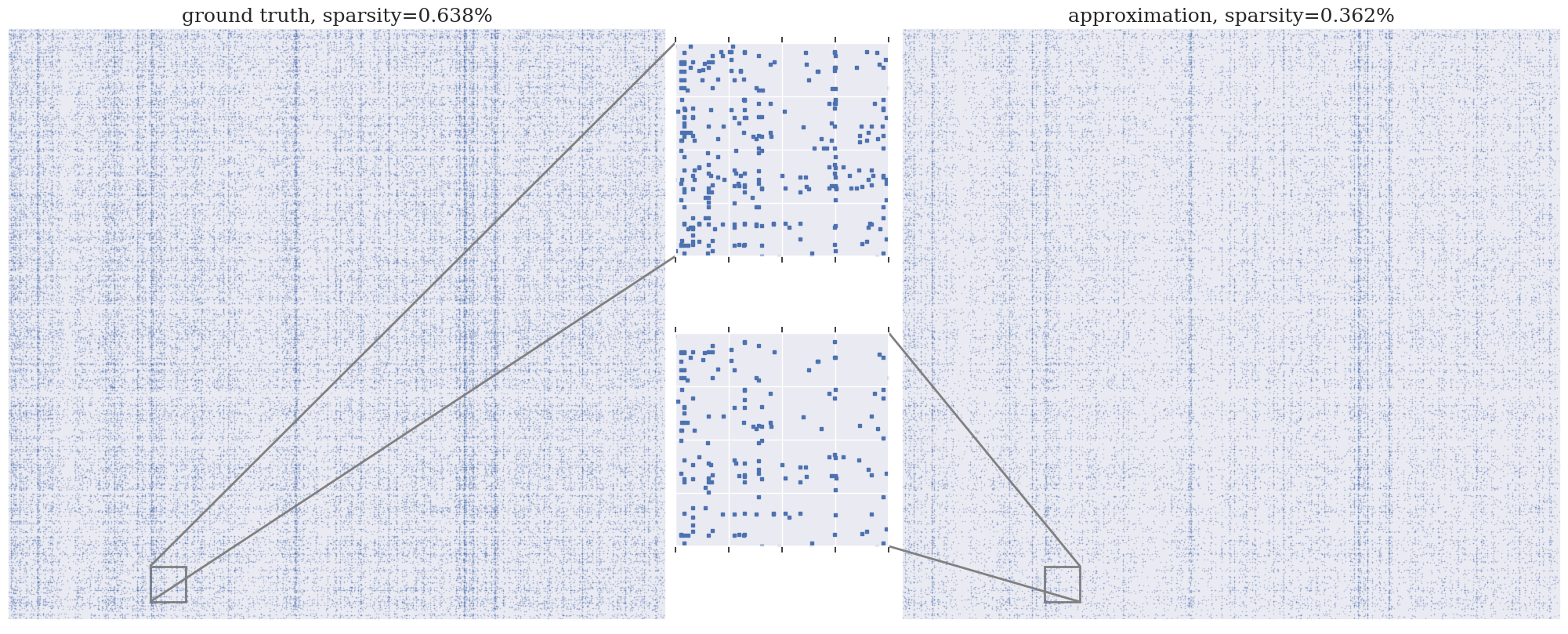}}{1}
    \note{-2.5,1.8}{ground truth, sparsity=$0.638\%$}
	\note{2.5,1.8}{approximation, sparsity=$0.362\%$}
	\end{annotate}
    \caption{Mask matrix visualization for \textbf{Citeseer}.}
    \label{fig:gae_citeseer}
\end{figure}

\begin{figure}[t]
    \centering
    \begin{annotate}{\includegraphics[width=\linewidth,trim={0 0 0 9mm},clip]{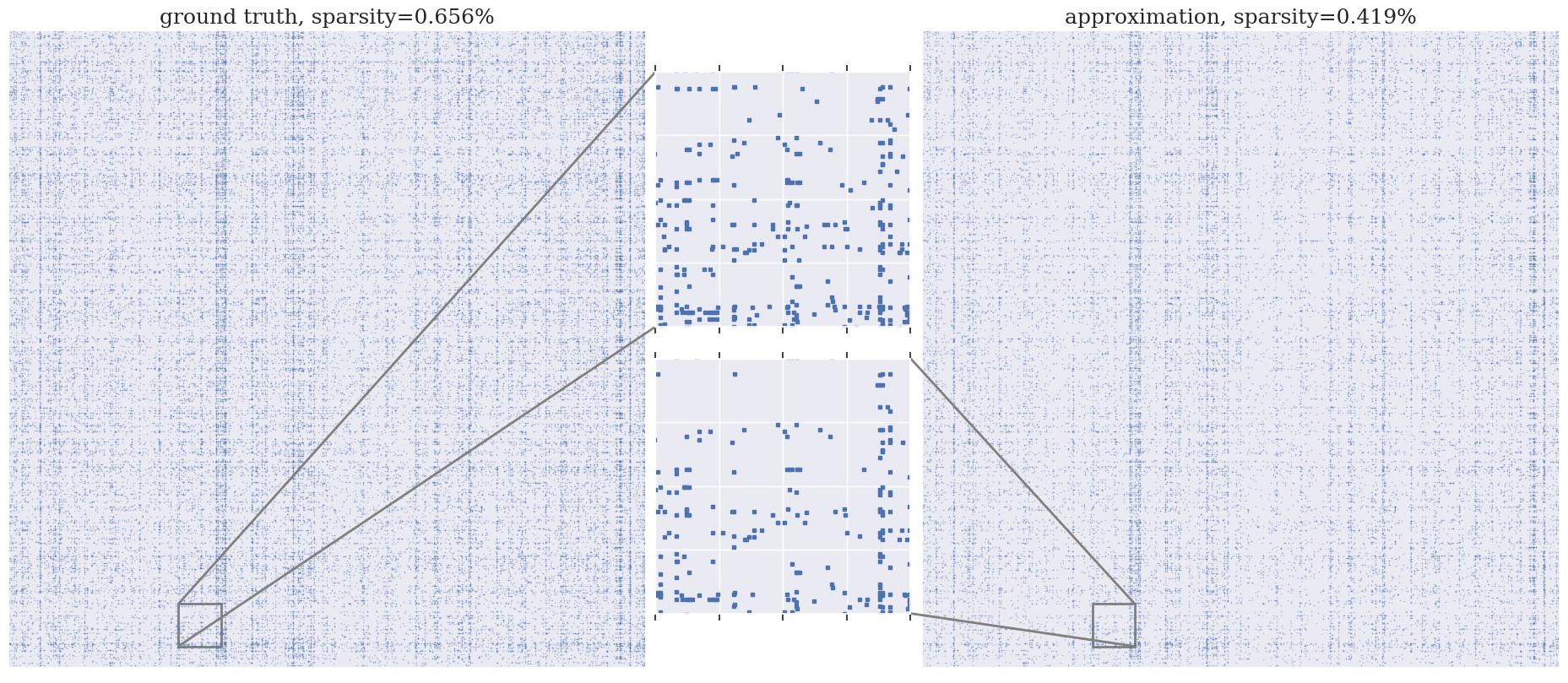}}{1}
    \note{-2.5,1.9}{ground truth, sparsity=$0.656\%$}
	\note{2.5,1.9}{approximation, sparsity=$0.419\%$}
	\end{annotate}
    \caption{Mask matrix visualization for \textbf{Coauthor-CS} with the first $3000$ nodes and $3000$ features.}
    \label{fig:gae_coauthor}
\end{figure}

\begin{figure}[t]
    \centering
    \includegraphics[width=0.8\linewidth]{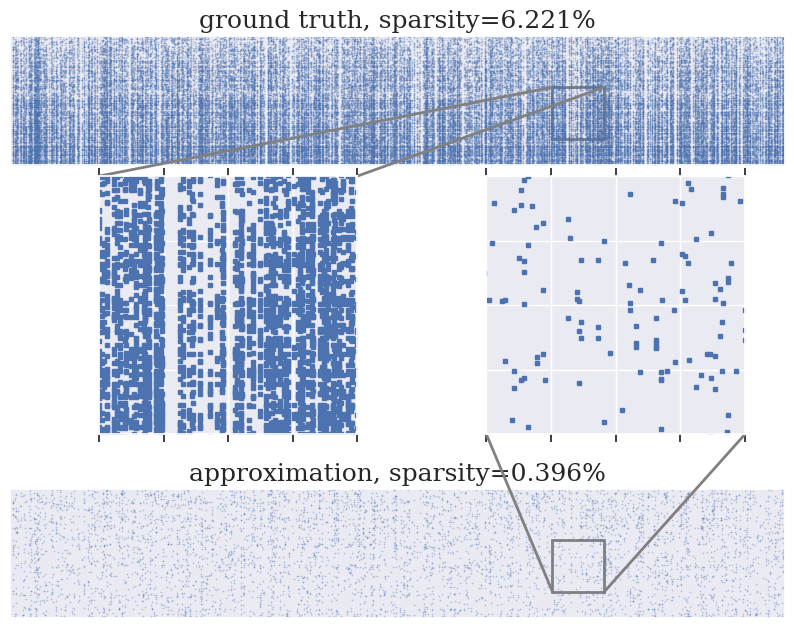}
    \caption{Mask matrix visualization for \textbf{PubMed} with the first $3000$ nodes and $500$ features at threshold=$0.005$.}
    \label{Fig:gae_pubmed}
\end{figure}

Figure~\ref{fig:gae_citeseer}-Figure~\ref{Fig:gae_pubmed} visualize the sparse mask matrix of the five datasets with anomaly injection. We are interested in the recall of the mask matrix that exposes the quality of the mask matrix approximation. As a result, we print the conditional mask matrix that marks the positions of approximated index that are at the true anomalous spots. Note that we did not visualize the mask matrix from the datasets under meta attack perturbation, as such attack focuses major poisonings on a minority of node entities. When making visualizations on such matrices, they are nothing more than a few horizontal lines. 
For \textbf{PubMed}, \textbf{Coauthor-CS} and \textbf{Wiki-CS}, we only visualize a subset of the full node attribute matrix, as the full matrix is too large to display. For instance, the size of \textbf{Coauthor-CS}'s node attribute matrix is about $\approx 18000\times 7000$. Different from the other three datasets, the attributes in \textbf{PubMed} and \textbf{Wiki-CS} has non-binary values. Consequently, we report the mask matrix with a threshold. In other words, in \textbf{Cora}, \textbf{Citeseer}, and \textbf{Coauthor-CS}, $M_{ij}$=1 if the $i,j$th value in the ground truth matrix and the perturbed matrix are different. In \textbf{PubMed}, $M_{ij}$=1 if the difference of the $j$th feature of the $i$th node in the ground truth matrix and the perturbed matrix are larger than $0.005$. The justification is similar to \textbf{Wiki-CS}, where the difference threshold is increased to $0.05$.


\begin{figure}[t]
    \centering
    \includegraphics[width=0.8\linewidth]{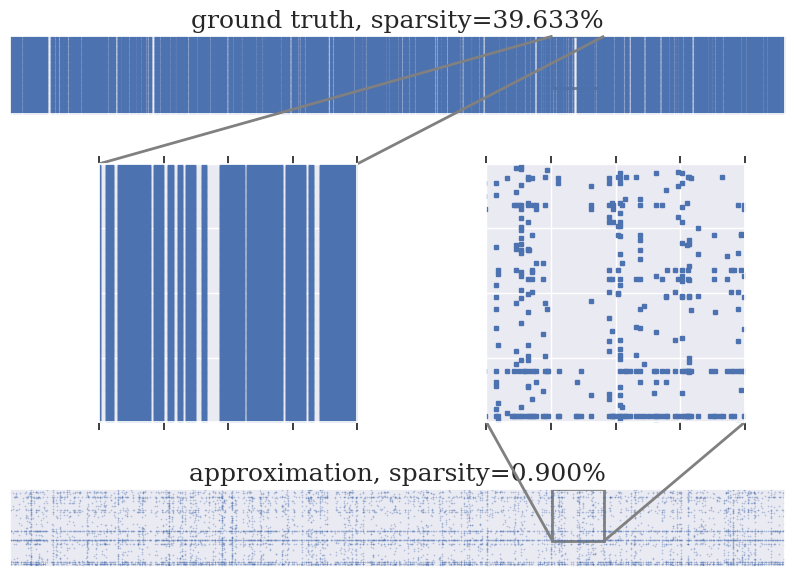}
    \caption{Mask matrix visualization for \textbf{Wiki-CS} with the first $3000$ nodes and $300$ features at threshold=$0.05$.}
    \label{fig:gae_wikics}
\end{figure}

\section{Reliability of Mask Approximation}
\label{sec:app:l1l2-tpfn}
\begin{figure*}[th]
    \centering
    \begin{minipage}{0.32\textwidth}
        \begin{annotate}
        {\includegraphics[width=\textwidth,trim={0 0 0 7mm},clip]{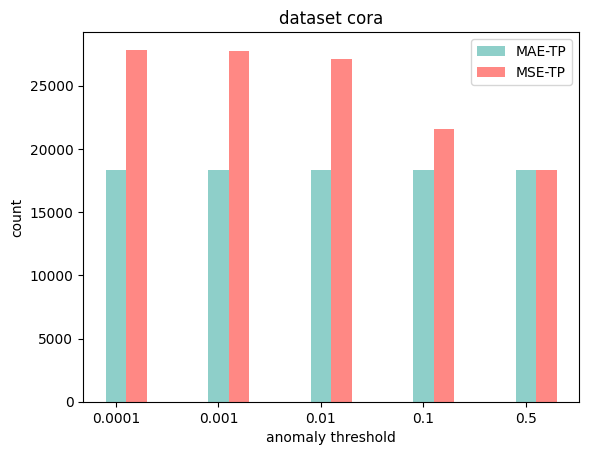}}{1}
        \note{0.2,2.2}{True Positive-Cora}
        \end{annotate}
    \end{minipage}
    \hspace{1mm}
    \begin{minipage}{0.32\textwidth}
        \begin{annotate}
        {\includegraphics[width=\textwidth,trim={0 0 0 7mm},clip]{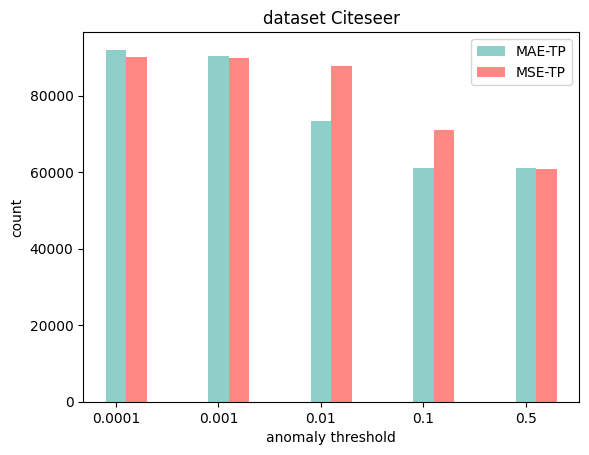}}{1}
        \note{0.2,2.2}{True Positive-CiteSeer}
        \end{annotate}
    \end{minipage}
    \hspace{1mm}
    \begin{minipage}{0.32\textwidth}
        \begin{annotate}
        {\includegraphics[width=\textwidth,trim={0 0 0 7mm},clip]{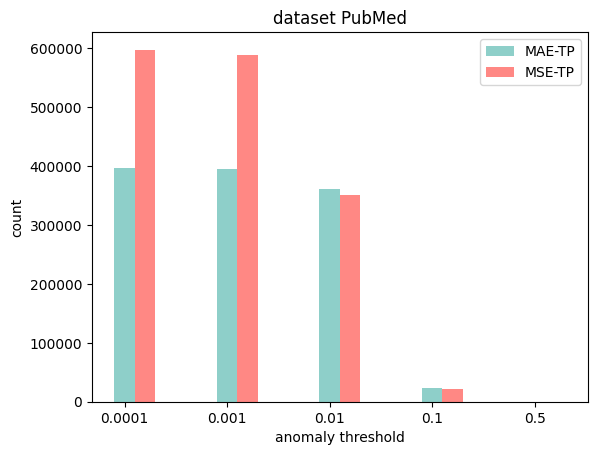}}{1}
        \note{0.2,2.2}{True Positive-PubMed}
        \end{annotate}
    \end{minipage}
    \begin{minipage}{0.32\textwidth}
        \begin{annotate}
        {\includegraphics[width=\textwidth,trim={0 0 0 7mm},clip]{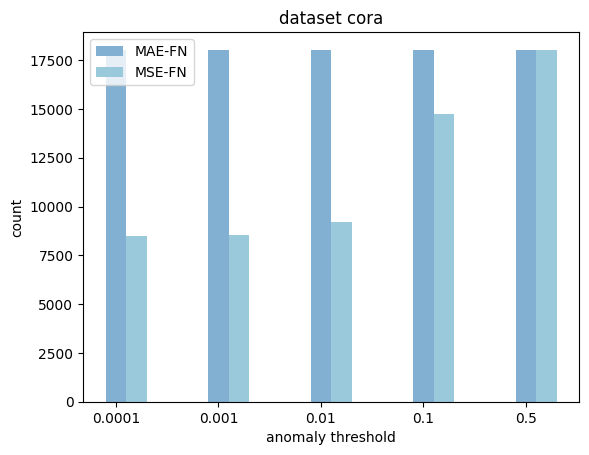}}{1}
        \note{0.2,2.2}{False Negative-Cora}
        \end{annotate}
    \end{minipage}
    \hspace{1mm}
    \begin{minipage}{0.32\textwidth}
        \begin{annotate}
        {\includegraphics[width=\textwidth,trim={0 0 0 7mm},clip]{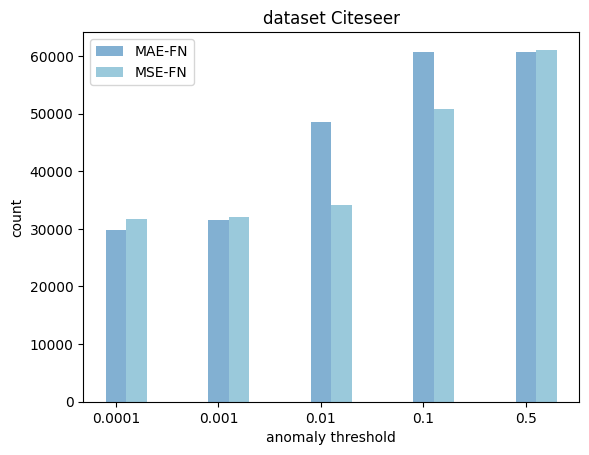}}{1}
        \note{0.2,2.2}{False Negative-CiteSeer}
        \end{annotate}
    \end{minipage}
    \hspace{1mm}
    \begin{minipage}{0.32\textwidth}
        \begin{annotate}
        {\includegraphics[width=\textwidth,trim={0 0 0 7mm},clip]{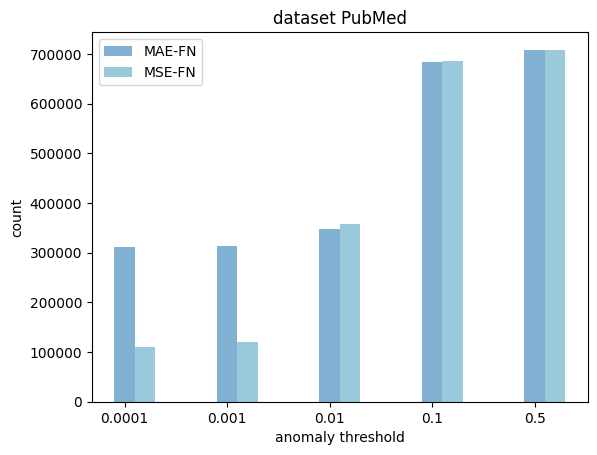}}{1}
        \note{0.2,2.2}{False Negative-PubMed}
        \end{annotate}
    \end{minipage}
    \begin{minipage}{0.32\textwidth}
        \begin{annotate}
        {\includegraphics[width=\textwidth,trim={0 0 0 8mm},clip]{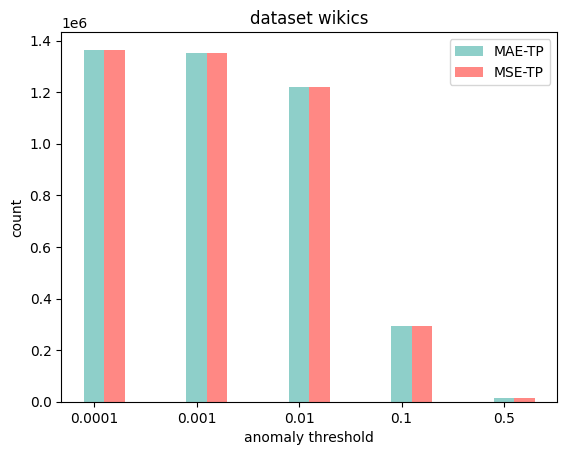}}{1}
        \note{0.2,2.3}{True Positive-WikiCS}
        \end{annotate}
    \end{minipage}
    \hspace{1mm}
    \begin{minipage}{0.32\textwidth}
        \begin{annotate}
        {\includegraphics[width=\textwidth,trim={0 0 0 7mm},clip]{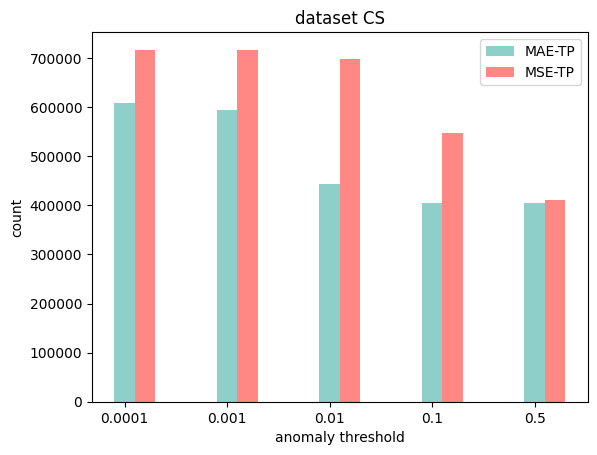}}{1}
        \note{0.2,2.2}{True Positive-CoauthorCS}
        \end{annotate}
    \end{minipage}
    \hspace{1mm}
    \begin{minipage}{0.32\textwidth}
        \begin{annotate}
        {\includegraphics[width=\textwidth,trim={0 0 0 7mm},clip]{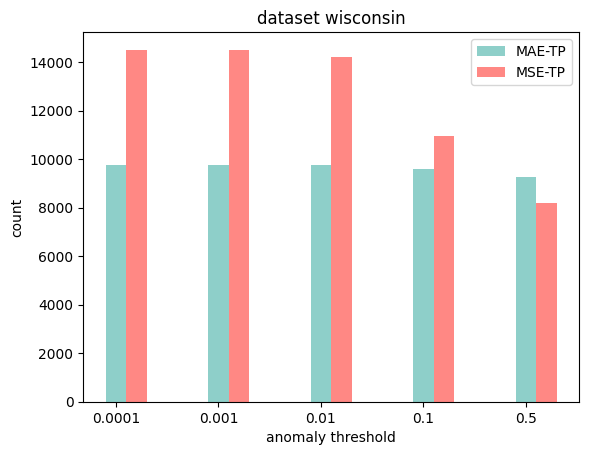}}{1}
        \note{0.2,2.2}{True Positive-Wisconsin}
        \end{annotate}
    \end{minipage}
    \begin{minipage}{0.32\textwidth}
        \begin{annotate}
        {\includegraphics[width=\textwidth,trim={0 0 0 7mm},clip]{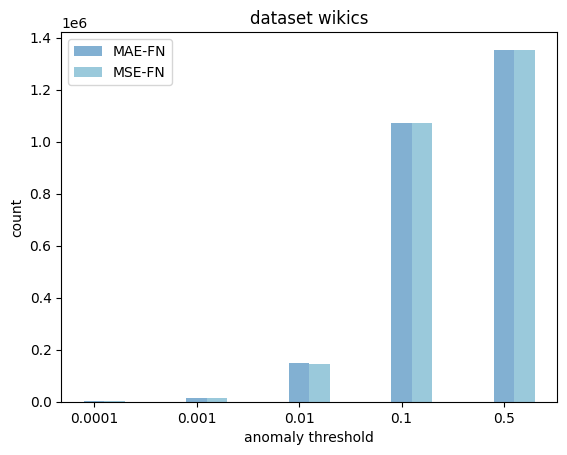}}{1}
        \note{0.2,2.2}{False Negative-WikiCS}
        \end{annotate}
    \end{minipage}
    \hspace{1mm}
    \begin{minipage}{0.32\textwidth}
        \begin{annotate}
        {\includegraphics[width=\textwidth,trim={0 0 0 7mm},clip]{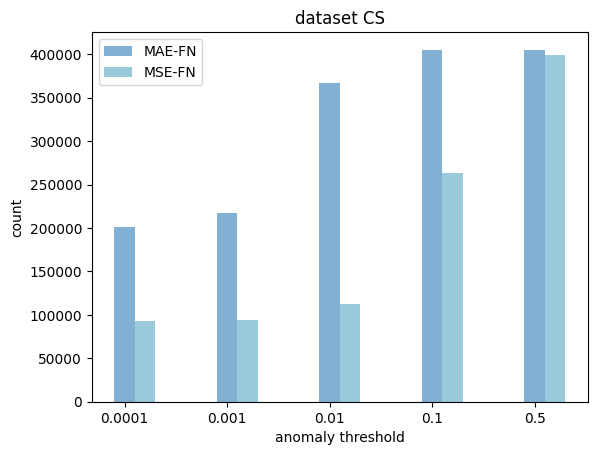}}{1}
        \note{0.2,2.2}{False Negative-CoauthorCS}
        \end{annotate}
    \end{minipage}
    \hspace{1mm}
    \begin{minipage}{0.32\textwidth}
        \begin{annotate}
        {\includegraphics[width=\textwidth,trim={0 0 0 7mm},clip]{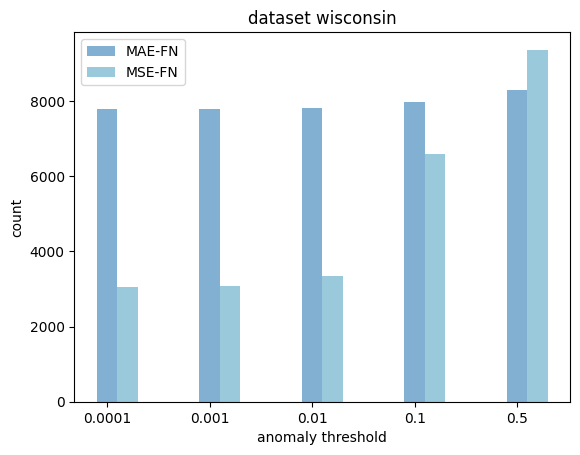}}{1}
        \note{0.2,2.2}{False Negative-Wisconsin}
        \end{annotate}
    \end{minipage}
    \begin{minipage}{0.32\textwidth}
        \begin{annotate}
        {\includegraphics[width=\textwidth,trim={0 0 0 7mm},clip]{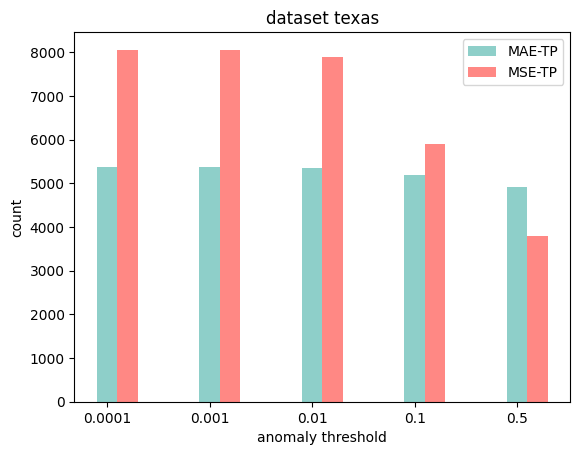}}{1}
        \note{0.2,2.2}{True Positive-Texas}
        \end{annotate}
    \end{minipage}
    \hspace{5mm}
    \begin{minipage}{0.32\textwidth}
        \begin{annotate}
        {\includegraphics[width=\textwidth,trim={0 0 0 7mm},clip]{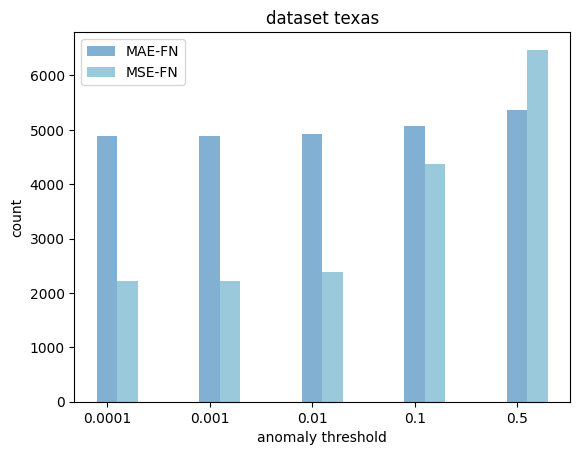}}{1}
        \note{0.2,2.2}{False Negative-Texas}
        \end{annotate}
    \end{minipage}
\caption{Comparison on the True Positive and False Negative of $\ell_1$ and $\ell_2$-based mask matrix generator on the seven datasets. The $\ell_1$-based results are named in the legend by MAE, and $\ell_2$-based results are denoted as MSE.}
\label{fig:tp_fn_mask}
\end{figure*}

We investigate the different choices on the regularizes of the mask matrix approximated by the GAE module. In the first ablation study of the main paper we compare the approximation results by $\ell_1$ and $\ell_2$ normalization in the loss function. Here we visualize the count of true positive and false negative predictions in Figure~\ref{fig:tp_fn_mask} for the seven (relatively) small datasets. We are more interested in identifying the non-zero positions, or positive approximations, as the majority of the mask matrix is filled with $0$.

It turns out that the $\ell_1$-based mask matrix does not change drastically among different threshold $\tau$s. In the contrast, tuning the threshold $\tau$ is critical for the $\ell_2$-based scoring function to perform a higher true positive rate and lower false negative rate. When the threshold is properly chosen, its approximation is more reliable than the $\ell_1$-based counterparts.

\end{document}